\documentclass[letterpaper]{article}
\usepackage[preprint]{aaai2027}
\usepackage[hyphens]{url}
\usepackage{graphicx}
\usepackage{amsmath,amssymb}
\usepackage{natbib}
\usepackage{caption}
\usepackage{algorithm}
\usepackage{algorithmic}
\usepackage{newfloat}
\usepackage{listings}
\DeclareCaptionStyle{ruled}{labelfont=normalfont,labelsep=colon,strut=off}
\floatstyle{ruled}
\newfloat{listing}{tb}{lst}{}
\floatname{listing}{Listing}

\usepackage{booktabs}
\usepackage{multirow}
\usepackage{array}
\usepackage{fontawesome5}

\newcommand{\projectpagelink}{%
    \leavevmode
    \pdfstartlink
    attr{/Border [0 0 0]}
    user{/Subtype /Link /A << /S /URI /URI (https://github.com/LouckXu/FedPAIE) >>}%
    \faGithub\hspace{0.35em}\textit{Project Page}%
    \pdfendlink
}

\title{Learning Color Grading, No Photo Sharing: Federated Aesthetic Preference Learning for Personalized Image Enhancement}
\author{
    Chuanzhi Xu\textsuperscript{\rm 1}\equalcontrib\corresponding,
    Ziyuan Tao\textsuperscript{\rm 2}\equalcontrib,
    Jean Julien KNell\textsuperscript{\rm 1},
    Yanrong Chen\textsuperscript{\rm 1},\\
    Haolan Guo\textsuperscript{\rm 1},
    Xuanhua Yin\textsuperscript{\rm 1},
    Adnan Mahmood\textsuperscript{\rm 2},
    Weidong Cai\textsuperscript{\rm 1}
}
\affiliations{
    \textsuperscript{\rm 1}The University of Sydney\qquad
    \textsuperscript{\rm 2}Macquarie University\\
    chuanzhi.xu@sydney.edu.au\qquad \projectpagelink
}

\begin{document}

\maketitle
\setlength{\parskip}{0pt}

% Non-anonymous arXiv main paper.
% Template-independent main-paper content.

% ---- Abstract ----
\begin{abstract}
Personalized image enhancement should reflect individual aesthetic taste, yet learning such preferences commonly depends on private photos and ratings that are unsuitable for centralized collection. The task must infer preference from sparse, heterogeneous feedback and translate it into natural-looking color transformations on resource-constrained user devices. We introduce FedPAIE, a federated personalized aesthetic image enhancement framework for user-adaptive color grading without centralizing raw photos or ratings. FedPAIE trains a lightweight dual-cue aesthetic scorer, calibrates it into a personalized scorer on a small local support set, and freezes it to guide regularized adaptation of a lightweight CLUT enhancer from unpaired local photographs. Fidelity constraints and an excess-gap penalty regularize scorer-guided adaptation to limit proxy-score over-optimization while preserving content and natural appearance. Training remains lightweight throughout the pipeline: scorer learning updates at most 0.787M parameters, enhancer adaptation updates 0.265M, and inference retains only a 0.293M-parameter personalized enhancer. Experiments on MIT-Adobe FiveK and Flickr-AES demonstrate effective open-world personalization and a favorable balance between user preference and image fidelity. FedPAIE thus connects decentralized preference learning with efficient personalized image transformation without requiring paired user retouches.
\end{abstract}

% ---- Introduction ----
\section{Introduction}
\label{sec:intro}

Personalized aesthetic image enhancement and color grading are common in digital photography, social media, and mobile content creation. Beyond exposure and contrast correction, users expect systems to adapt to preferences for color temperature, saturation, tone curves, and stylistic mood. Platforms could learn these preferences from private photos, ratings, editing histories, or text prompts, but such sensitive data are unsuitable for centralized collection. This raises a question: \emph{how can an aesthetic enhancement model learn personalized color-grading preferences while keeping user data private?}

\begin{figure}
    \centering
    \includegraphics[width=\linewidth]{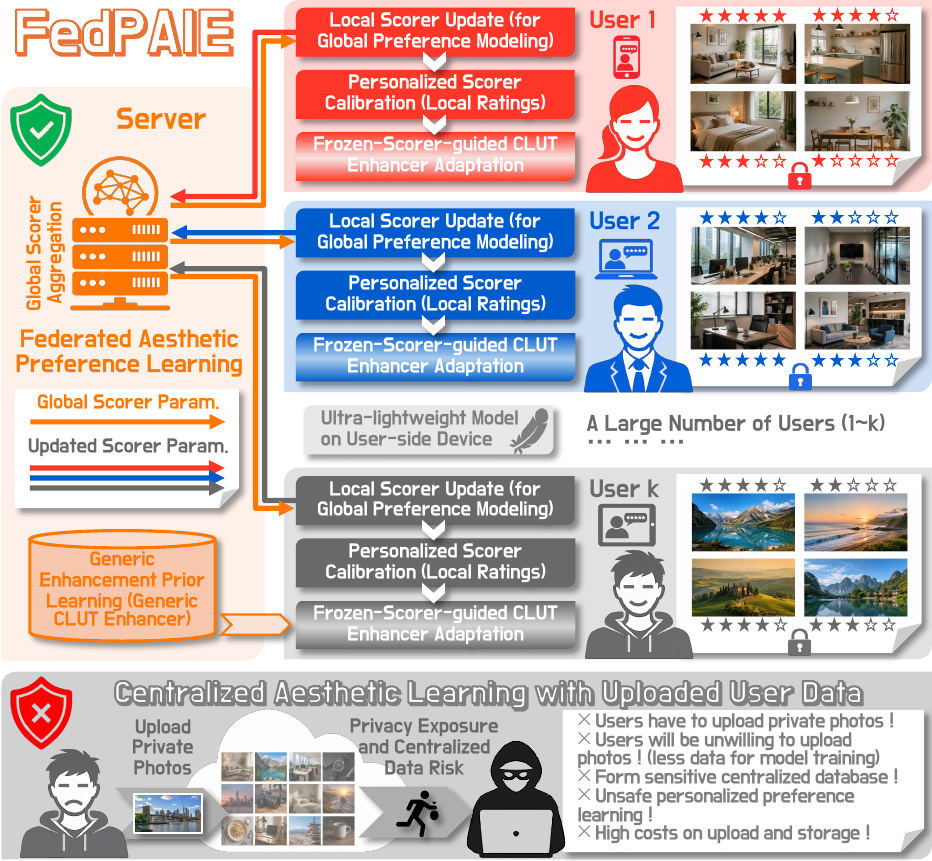}
    \caption{FedPAIE learns user-specific color grading without centralizing
    private photos or ratings. Federated Aesthetic Preference Learning yields
    a global scorer, locally calibrated into a personalized scorer and frozen
    to guide a lightweight 0.293M-parameter CLUT enhancer on unpaired user
    images.}
    \label{fig:teaser}
\end{figure}

Existing image enhancement methods learn efficient, image-adaptive color and tone transformations from paired data~\cite{bychkovsky2011learning,zhang2022clutnet,kim2025svdlut}. Meanwhile, personalized enhancement methods model user-specific retouching preferences~\cite{kim2020pienet,bianco2020personalized,kosugi2024personalized}, while personalized aesthetic assessment predicts individual deviations from population-level aesthetics~\cite{ren2017personalized}. However, many existing approaches still assume centralized access to user data, paired user retouching examples, or preference annotations that may be sensitive in real consumer applications. These assumptions are misaligned with privacy-sensitive image-editing scenarios, where users may provide only limited feedback and may refuse to share raw photos or preference prompts, as illustrated in the lower panel of Fig.~\ref{fig:teaser}.

Federated learning offers a data-local approach to preference learning because it trains models from decentralized user data without directly uploading raw samples to a central server~\cite{mcmahan2017communication,kairouz2021advances}. Recent work has demonstrated federated visual learning in deployed object detection, general computer-vision benchmarks, personalized aesthetic assessment, and parameter-efficient video moderation~\cite{liu2020fedvision,he2021fedcv,xiong2023federated,tao2025fedvideomae}.

However, applying federated learning to personalized aesthetic enhancement remains challenging. First, federated personalization requires client-side model execution, yet many aesthetic enhancement models are too heavy for user devices~\cite{kim2020pienet,kosugi2024personalized}. User preferences are highly subjective, while an individual user typically provides explicit ratings for only a small number of images, leaving each client with limited supervision for learning personal taste. Meanwhile, rating scales and aesthetic distributions vary substantially across users, yielding non-independent-and-identically-distributed (non-IID) federated data. Standard federated aggregation can learn a shared initialization from these decentralized signals but may not fully capture every user's aesthetic preferences. Moreover, personalized aesthetic enhancement introduces an additional challenge: the learned preference model must be used as a training signal for image transformation. Since lightweight aesthetic scorers provide only weak and imperfect supervision, directly optimizing an enhancer against their predictions may exploit imperfections in the proxy objective~\cite{amodei2016concrete}, motivating explicit fidelity regularization. Therefore, a practical model must balance privacy preservation, user-specific adaptation, computational efficiency, and optimization stability.

In this paper, we propose \textbf{FedPAIE} (\textbf{Fed}erated \textbf{P}ersonalized \textbf{A}esthetic \textbf{I}mage \textbf{E}nhancement), to our knowledge the first federated method for personalized aesthetic image enhancement and color grading. As illustrated in Fig.~\ref{fig:teaser}, FedPAIE separates Federated Aesthetic Preference Learning and Generic Enhancement Prior Learning from private On-Device Preference Adaptation, keeping raw photos and ratings local while each client stage trains at most 0.787M parameters and inference retains only a 0.293M-parameter enhancer. Our contributions are summarized as follows:
\begin{itemize}
    \item We introduce FedPAIE, to our knowledge the first federated method for personalized aesthetic image enhancement and color grading, enabling unpaired enhancement while keeping user data local.

    \item We develop the Lightweight Dual-Cue Aesthetic Scorer, Federated Aesthetic Preference Learning, Generic Enhancement Prior Learning, and Personalized Scorer Calibration. We further introduce the regularized Personalized Enhancement Objective and checkpoint selection criterion for Frozen-Scorer-Guided Enhancer Adaptation.

    \item Extensive experiments demonstrate effective open-world personalization and a favorable preference--fidelity trade-off. Five matched ablation settings validate the roles of key objective terms and functional groups.
\end{itemize}

\begin{figure*}
    \centering
    \includegraphics[width=\textwidth]{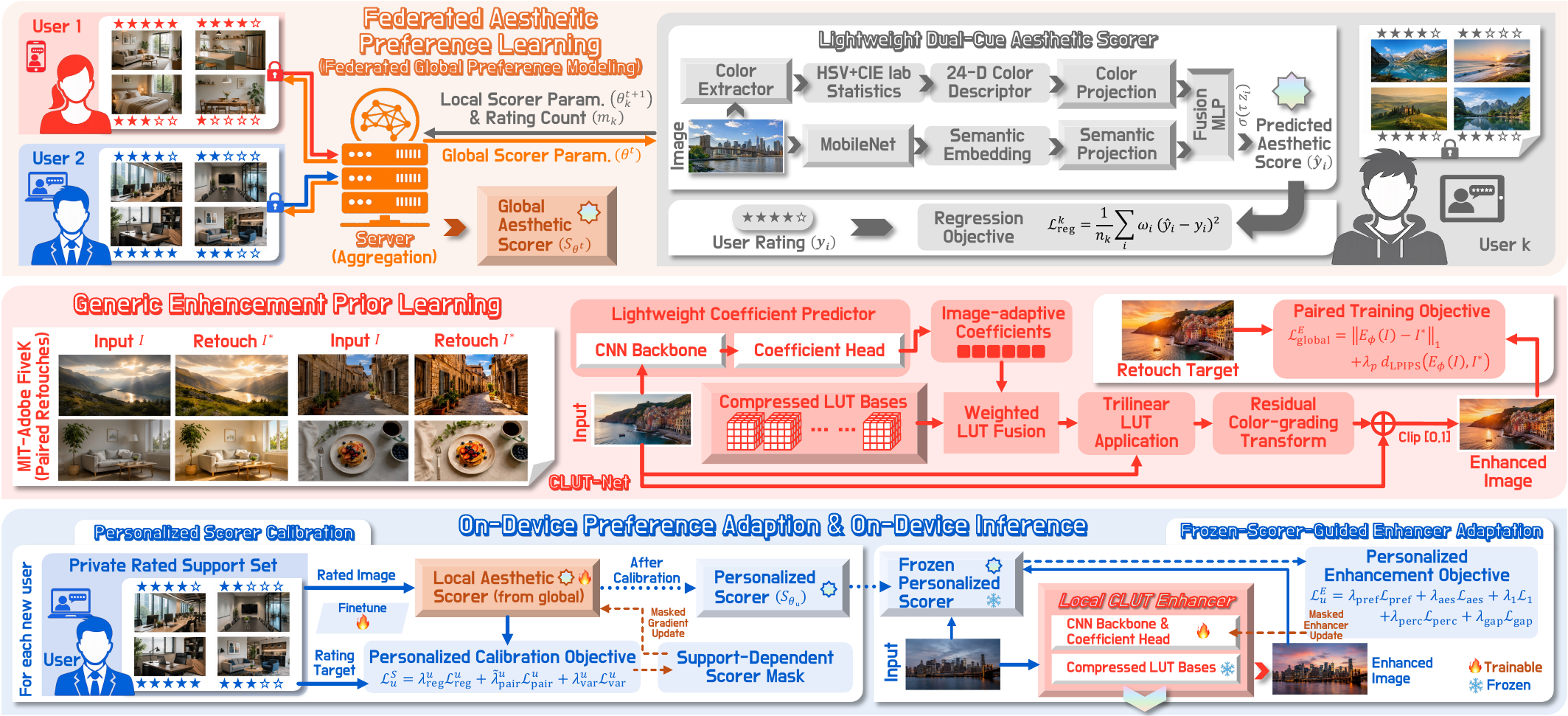}
    \caption{Overview of FedPAIE. Global initialization learns the Lightweight Dual-Cue Aesthetic Scorer and CLUT enhancer through Federated Aesthetic Preference Learning and Generic Enhancement Prior Learning. On-Device Preference Adaptation performs Personalized Scorer Calibration followed by Frozen-Scorer-Guided Enhancer Adaptation. Personalized On-Device Inference retains only the lightweight personalized enhancer.}
    \label{fig:fedpaie_framework}
\end{figure*}

% ---- Related Work ----
\section{Related Work}

We briefly review related work here, with complete analysis and discussion in
Appendix Sec.~I.
\par\noindent\textbf{Personalized Aesthetics Assessment and Enhancement.}
Generic aesthetic modeling spans population-level image assessment, aesthetics-aware diffusion generation, 3D scene assessment, etc.~\cite{ke2023vila,yin2026accelaes,xu2026aes3d}, while personalized assessment methods predict user-specific judgments from attributes, few-shot adaptation, graph collaboration, transitional contrast learning, task-vector customization, or continual feedback~\cite{yang2022personalized,zhu2022personalized,shi2024personalized,yang2024multilevel,yun2024scaling,zhong2025paaplus}. These personalized assessment methods output scores rather than personalized transformations. Personalized enhancement learns user-specific retouching through preference embeddings, neural-spline transforms, masked style modeling, or global-local style conditioning~\cite{kim2020pienet,bianco2020personalized,kosugi2024personalized,kim2025global}. Recent systems infer photographic styles from pairwise judgments or combine VLM-driven interaction and scene-aware memory with semantic retouching~\cite{kim2026pps,chang2026pertouch}. FedPAIE differs in learning from federated sparse scalar ratings, keeping raw photos and ratings local, and using a calibrated scorer for unpaired on-device enhancement without user-specific retouch targets.
\par\noindent\textbf{Federated Preference Learning.}
Federated learning and non-IID variants keep raw samples local during training~\cite{mcmahan2017communication,li2020federated,mohri2019agnostic}. Federated recommenders learn private preferences~\cite{ammad2019federated,chai2021secure,liang2021fedrec,yi2021efficient,liu2023federated}, while federated personalized image-aesthetics assessment predicts user-specific scores~\cite{xiong2023federated}. To our knowledge, FedPAIE is the first federated method for personalized aesthetic image enhancement and color grading.

% ---- Methodology ----
\section{Methodology -- FedPAIE}
\label{sec:method}

\subsection{Framework Overview}

FedPAIE realizes personalized color grading with an image-adaptive 3D LUT through Generic Enhancement Prior Learning and private On-Device Preference Adaptation. As shown in Fig.~\ref{fig:fedpaie_framework}, the pipeline has three stages. Global initialization independently learns a population-level scorer through Federated Aesthetic Preference Learning and a CLUT enhancer through Generic Enhancement Prior Learning. On-device adaptation calibrates the scorer on a private rated support set, freezes it, and then adapts the enhancer on unpaired local photographs. Personalized On-Device Inference removes the scorer and retains only the lightweight personalized enhancer. Raw photographs and ratings remain local, and no paired user-specific retouches are required. End-to-end pseudocode, masked parameter updates, and the component lifecycle appear in Appendix Secs.~A, B, and D.

Formally, client $k$ owns private rated samples $\mathcal{D}_k=\{(I_i,y_i)\}_{i=1}^{n_k}$, where $I_i$ is the $i$-th image, $y_i\in[0,1]$ is its normalized aesthetic rating, and $n_k=|\mathcal D_k|$. Let $\theta^g$ and $\phi^g$ denote the global scorer and generic enhancer parameters, yielding $S_{\theta^g}$ and $E_{\phi^g}$. For a new user $u$, FedPAIE uses a rated support set $\mathcal D_u^s$ and unpaired local images $\mathcal U_u$ to obtain personalized parameters $\theta_u$ and $\phi_u$. The output $\hat I=E_{\phi_u}(I)$ should receive a higher user-specific preference score than $I$ while preserving its content and natural appearance.

\subsection{Federated Aesthetic Preference Learning}
\par\noindent\textbf{Lightweight Dual-Cue Aesthetic Scorer.}
The scorer combines low-level color statistics with high-level semantic context. For image $I_i$, the fixed color extractor $\Phi_c$ computes the mean, standard deviation, minimum, and maximum of each HSV and CIE Lab channel, yielding a 24-dimensional descriptor $\mathbf{c}_i=\Phi_c(I_i)$. A shared, fixed ImageNet-pretrained MobileNetV3-Large~\cite{howard2019searching} serves as the semantic extractor $\Phi_s$ and produces a 960-dimensional embedding $\mathbf{s}_i=\Phi_s(I_i)$. Both feature extractors are reused by all clients and remain frozen throughout global and personalized scorer training. Trainable projections $P_c$ and $P_s$, parameterized by $\theta_c$ and $\theta_s$, map the color and semantic cues to 512- and 256-dimensional latent representations:
\begin{equation}
    \mathbf{h}_i^c=P_c(\mathbf{c}_i),\qquad
    \mathbf{h}_i^s=P_s(\mathbf{s}_i).
\end{equation}
The two latent representations are concatenated and processed by a fusion MLP $G_{\theta_f}$ parameterized by $\theta_f$. Together with a learnable temperature $\tau$, the trainable scorer parameters are $\theta=(\theta_c,\theta_s,\theta_f,\tau)$. The unconstrained scalar output is mapped to the interval $(0,1)$ as:
\begin{equation}
    \hat y_i=S_\theta(I_i)
    =\sigma\!\left(\tau\,G_{\theta_f}
    \bigl([\mathbf{h}_i^c\Vert\mathbf{h}_i^s]\bigr)\right),
    \label{eq:scorer}
\end{equation}
where $\tau$ is constrained to a fixed interval and $\sigma$ is the sigmoid function. Thus, $\hat y_i\in(0,1)$ is predicted on the same normalized scale as $y_i\in[0,1]$. The dual-cue design keeps the trainable scorer lightweight while retaining the color sensitivity needed for grading and the semantic context needed to assess whether a transformation suits the image.
\par\noindent\textbf{Federated Global Preference Modeling.}
The federated stage learns a population-level scorer initialization using rating regression only. Pairwise ordering and variance preservation are introduced during user-specific calibration. For client $k$, the regression objective is:
\begin{equation}
    \mathcal{L}_{\mathrm{reg}}^k
    =\frac{1}{n_k}\sum_i \omega_i(\hat y_i-y_i)^2,
    \label{eq:scorer_reg}
\end{equation}
where $\omega_i=1$ by default. When a client's rating distribution exceeds a prespecified imbalance threshold, $\omega_i$ becomes the normalized and clipped inverse frequency of the rating bin containing $y_i$. This optional reweighting changes individual error contributions without adding an objective. Appendix Sec.~B gives the binning, activation, normalization, and clipping rules. The global client objective is therefore:
\begin{equation}
    \mathcal{L}_{\mathrm{FL}}^k
    =\mathcal{L}_{\mathrm{reg}}^k.
    \label{eq:fl_scorer_loss}
\end{equation}
Let $\theta^t$ denote the server-side scorer parameters at the start of communication round $t$, with $\theta^0$ denoting their initialization, and let $\mathcal C_t$ be the participating clients. The server sends $\theta^t$ to each $k\in\mathcal C_t$. After minimizing Equation~\eqref{eq:fl_scorer_loss}, client $k$ returns only its updated parameters $\theta_k^{t+1}$ and the count $m_k$ of examples processed across its local steps. No raw sample or rating is transmitted. We use a square-root-reweighted variant of FedAvg~\cite{mcmahan2017communication}:
\begin{equation}
    \theta^{t+1}=\sum_{k\in\mathcal{C}_t}\alpha_k\theta_k^{t+1},
    \qquad
    \alpha_k=\frac{\sqrt{m_k}}{\sum_{j\in\mathcal{C}_t}\sqrt{m_j}}.
    \label{eq:fedavg}
\end{equation}
Here $\alpha_k$ is client $k$'s normalized aggregation weight. Square-root weighting retains a notion of client evidence while reducing domination by users with many more ratings. FedPAIE establishes protocol-level raw-data locality: photographs and ratings remain on client devices, while federated communication contains only lightweight scorer parameters and aggregation counts. Secure aggregation and differential privacy are compatible communication-layer extensions, as detailed in Appendix Secs.~D and H.

\begin{figure*}[t]
\centering
\includegraphics[width=\textwidth]{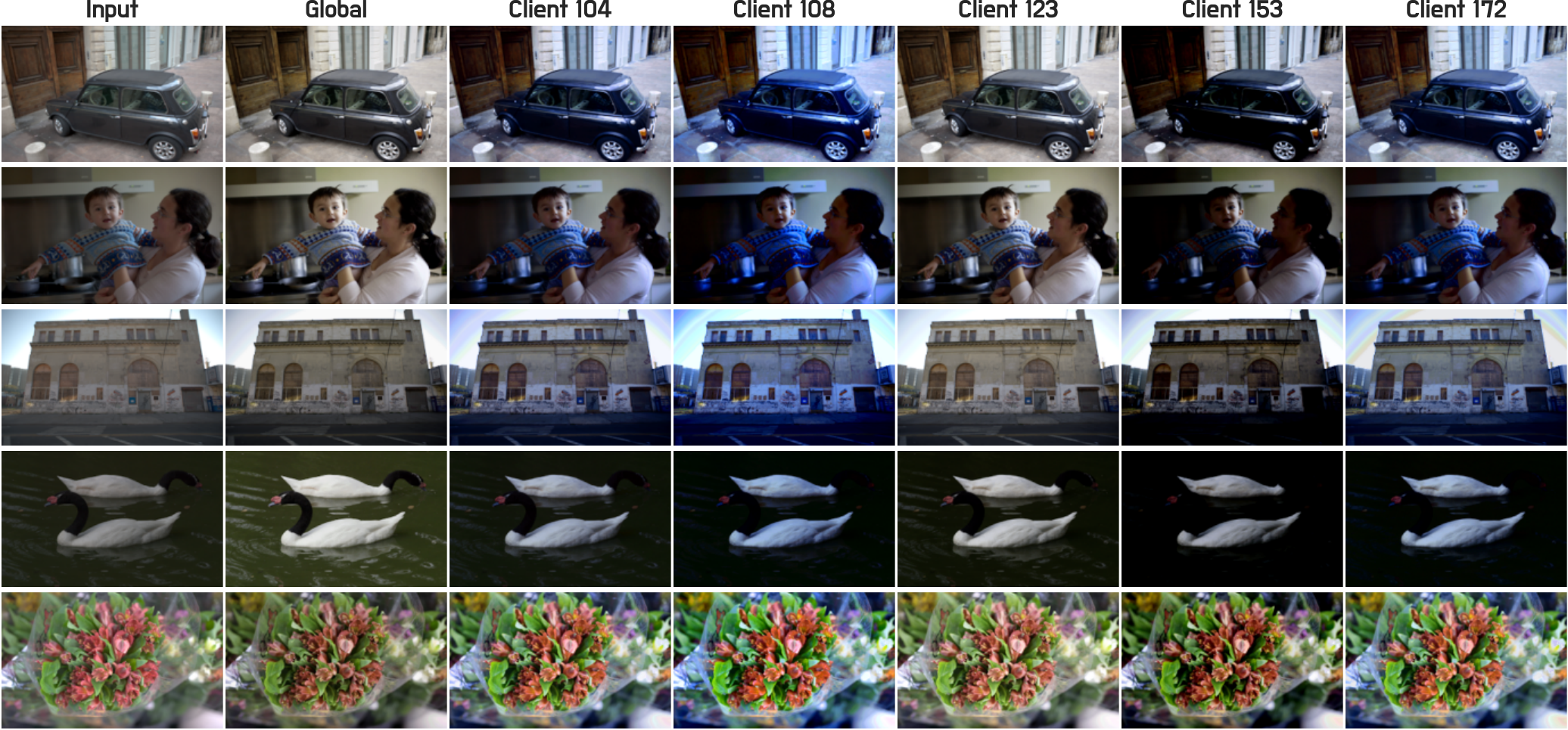}
\caption{Five shared-input FiveK comparisons. The CLUT enhancer obtained through
Generic Enhancement Prior Learning and five personalized enhancers produce distinct temperature, exposure, and
contrast while preserving scene structure. Appendix Sec.~G provides detail crops
and full-cohort analyses.}
\label{fig:qualitative_main}
\end{figure*}

\subsection{Generic Enhancement Prior Learning}

We use CLUT-Net~\cite{zhang2022clutnet} as the generic enhancer because it represents color grading as an efficient image-adaptive transform. Let $\phi$ denote its complete parameter set, including a lightweight coefficient predictor and compressed LUT bases. Given $I$, the coefficient predictor $W_{\phi}$, comprising a CNN backbone and coefficient head, produces image-adaptive coefficients $\mathbf{w}(I)=[w_1(I),\ldots,w_M(I)]$. Weighted fusion combines $M$ compressed LUT bases $\{\mathbf B_q\}_{q=1}^M$ into an image-specific residual LUT:
\begin{equation}
    \mathbf{L}_{\phi}(I)=\sum_{q=1}^{M}w_q(I)\mathbf{B}_q.
    \label{eq:lut_fusion}
\end{equation}
The compressed bases are reconstructed from factorized parameters. Trilinear LUT application $\mathcal T$ evaluates the fused LUT at the input RGB values to produce a residual color-grading transform. Appendix Sec.~C gives the factor dimensions, basis reconstruction, and frozen-basis personalization rule. Adding this residual to the input gives the unclipped output:
\begin{equation}
    \widetilde E_\phi(I)
    =I+\mathcal{T}\bigl(I,\mathbf{L}_{\phi}(I)\bigr).
    \label{eq:clut_raw_output}
\end{equation}
During Frozen-Scorer-Guided Enhancer Adaptation and Personalized On-Device Inference, an external clipping step produces the valid image:
\begin{equation}
    E_\phi(I)=\operatorname{clip}\!\left(\widetilde E_\phi(I),0,1\right).
    \label{eq:clut_output}
\end{equation}
The residual formulation preserves the input as a natural reference and enables full-resolution enhancement without a heavy pixel-generating decoder.

FedPAIE initializes $\phi^g$ from a CLUT-Net checkpoint pretrained on paired MIT-Adobe FiveK retouching data. For a pair $(I,I^*)$, where $I^*$ is an expert-retouched target, the Paired Training Objective evaluates pixel fidelity and LPIPS~\cite{zhang2018unreasonable} on the unclipped output $\widetilde E_\phi(I)$:
\begin{equation}
\begin{split}
    \mathcal{L}_{\mathrm{global}}^{E}
    ={}&\mathbb{E}_{(I,I^*)}
    \left[\|\widetilde E_\phi(I)-I^*\|_1\right]\\
    &+\lambda_p\mathbb{E}_{(I,I^*)}
    \left[d_{\mathrm{LPIPS}}(\widetilde E_\phi(I),I^*)\right].
\end{split}
    \label{eq:global_enhancer}
\end{equation}
Here $\lambda_p\geq0$ weights the perceptual term. This initialization supplies diverse, generally useful color transformations before any private preference signal is introduced.

\subsection{On-Device Preference Adaptation}
\par\noindent\textbf{Personalized Scorer Calibration.}
For user $u$, we initialize the local scorer from $\theta^g$ and apply the Support-Dependent Scorer Mask $\mathbf M_u^S$ to select the parameter blocks adapted on the private rated support set $\mathcal D_u^s$. The 10-shot regime updates only $(\theta_f,\tau)$. The 100-shot regime also updates $(\theta_c,\theta_s)$, while both feature extractors remain fixed. This policy restricts capacity under sparse supervision and permits stronger feature alignment when more ratings are available. Appendix Sec.~B gives the exact binary masks and their relation to the implementation cutoff.
Thus, 10-shot calibration adjusts the fusion and rating scale without relearning cue projections, while 100-shot calibration can realign both projected cues. This explicitly controls adaptation capacity instead of fine-tuning the full scorer from sparse ratings.

Beyond the global regression objective, local calibration models relative preferences and discourages prediction collapse. For a local mini-batch, let $\delta>0$ be the minimum normalized-rating separation, let $\mathcal{P}_u=\{(i,j):i<j,\ |y_i-y_j|>\delta\}$ contain the resulting unordered pairs, and let $r_{ij}=\operatorname{sign}(y_i-y_j)$. We use the smooth pairwise objective:
\begin{equation}
    \mathcal{L}_{\mathrm{pair}}^u
    =-\frac{1}{|\mathcal{P}_u|}
    \sum_{(i,j)\in\mathcal{P}_u}
    \log\sigma\!\left(r_{ij}(\hat y_i-\hat y_j)\right).
    \label{eq:personal_rank}
\end{equation}
Let $\mathbf y$ and $\hat{\mathbf y}$ collect the target and predicted ratings in the same mini-batch. We also use the batchwise variance-preservation term:
\begin{equation}
    \mathcal{L}_{\mathrm{var}}^u
    =\left[\rho\,\operatorname{Std}(\mathbf{y})
    -\operatorname{Std}(\hat{\mathbf{y}})\right]_+,
    \qquad [z]_+=\max(z,0),
    \label{eq:personal_var}
\end{equation}
where $\rho\in(0,1]$ specifies the fraction of target-score dispersion to preserve. Let $\mathcal L_{\mathrm{reg}}^u$ denote Equation~\eqref{eq:scorer_reg} evaluated on $\mathcal D_u^s$. The weighted regression, pairwise, and variance terms form the personalized calibration objective $\mathcal L_u^S$ in Fig.~\ref{fig:fedpaie_framework}. We collect the adaptable scorer parameters in $\vartheta$, with feasible set $\mathcal A_u$, and hold all remaining parameters at their global values $\theta_{\mathrm{fix}}^g$. Calibration solves:
\begin{equation}
\begin{split}
    \vartheta_u=\arg\min_{\vartheta\in\mathcal{A}_u}
    \left(
    \lambda_{\mathrm{reg}}^u\mathcal{L}_{\mathrm{reg}}^u
    +\widetilde{\lambda}_{\mathrm{pair}}^u
    \mathcal{L}_{\mathrm{pair}}^u
    +\lambda_{\mathrm{var}}^u\mathcal{L}_{\mathrm{var}}^u
    \right),\\
    \theta_u=(\theta_{\mathrm{fix}}^g,\vartheta_u).
\end{split}
    \label{eq:personal_scorer}
\end{equation}
The nonnegative coefficients $\lambda_{\mathrm{reg}}^u$, $\widetilde{\lambda}_{\mathrm{pair}}^u$, and $\lambda_{\mathrm{var}}^u$ weight regression, pairwise ordering, and variance preservation. The effective pairwise coefficient incorporates support-regime selection and optional collapse protection, as specified in Appendix Sec.~B. If the support set does not contain reliable ordered pairs, the pairwise term is omitted. After calibration, $S_{\theta_u}$ is frozen and used only as a training-time preference model.
\par\noindent\textbf{Frozen-Scorer-Guided Enhancer Adaptation.}
We initialize a local CLUT enhancer from $\phi^g$ and freeze both its compressed LUT bases and the personalized scorer. Only the CNN backbone and coefficient head of the lightweight coefficient predictor are updated. Freezing the bases preserves the transformation dictionary learned from paired retouches, while updating the predictor personalizes how those transformations are mixed rather than learning unconstrained LUTs from sparse unpaired data. For an unpaired local image $I\in\mathcal{U}_u$, let $\hat I=E_\phi(I)$ and define:
\begin{equation}
    s_u^+=S_{\theta_u}(\hat I),\qquad
    s_u^0=S_{\theta_u}(I),\qquad
    \Delta_u=s_u^+-s_u^0.
    \label{eq:score_gap}
\end{equation}
The scorer parameters remain fixed, but gradients propagate through $S_{\theta_u}$ to the enhancer. This separates a stable preference objective from the transformation being optimized and prevents joint scorer--enhancer drift. Appendix Sec.~B gives the masked update and gradient derivation. All expectations below are over $I\sim\mathcal U_u$. The preference terms are:
\begin{equation}
    \mathcal{L}_{\mathrm{pref}}=-\mathbb{E}_{I}\log\sigma(\Delta_u),
    \qquad
    \mathcal{L}_{\mathrm{aes}}=-\mathbb{E}_{I}s_u^+.
    \label{eq:preference_losses}
\end{equation}
The fidelity terms are:
\begin{equation}
    \mathcal{L}_{1}=\mathbb{E}_{I}\|\hat I-I\|_1,
    \qquad
    \mathcal{L}_{\mathrm{perc}}
    =\mathbb{E}_{I}d_{\mathrm{LPIPS}}(\hat I,I).
    \label{eq:fidelity_losses}
\end{equation}
The excess-gap penalty is:
\begin{equation}
    \mathcal{L}_{\mathrm{gap}}
    =\mathbb{E}_{I}\left[\Delta_u-\mu\right]_+,
    \label{eq:gap_loss}
\end{equation}
where $\mu\geq0$ is the tolerated gain. This term discourages scorer-proxy exploitation. The Personalized Enhancement Objective is:
\begin{equation}
\begin{split}
    \mathcal{L}_{u}^{E}
    ={}&\lambda_{\mathrm{pref}}\mathcal{L}_{\mathrm{pref}}
    +\lambda_{\mathrm{aes}}\mathcal{L}_{\mathrm{aes}}
    +\lambda_{1}\mathcal{L}_{1}\\
    &+\lambda_{\mathrm{perc}}\mathcal{L}_{\mathrm{perc}}
    +\lambda_{\mathrm{gap}}\mathcal{L}_{\mathrm{gap}}.
\end{split}
    \label{eq:personal_enhancer}
\end{equation}
The nonnegative weights balance relative and absolute preference improvement, image fidelity, and proxy-exploitation control. The preference and aesthetic terms encourage relative and absolute improvement, the pixel and perceptual terms preserve content and natural appearance, and the excess-gap penalty limits exploitation of an imperfect scorer. Training minimizes Equation~\eqref{eq:personal_enhancer} over the feasible parameter set $\Phi_u$, which contains only the coefficient predictor's CNN backbone and head while keeping the LUT bases fixed.

The training objective regularizes individual outputs, while model selection controls the checkpoint-level preference--fidelity trade-off. To avoid selecting a checkpoint solely for a high proxy score, let $\mathcal V_u$ be the local validation subset and $\mathcal H_u$ the candidate checkpoint set. We score each $\phi\in\mathcal H_u$ as:
\begin{equation}
\begin{split}
    Q_u(\phi)={}&\mathbb{E}_{I\in\mathcal{V}_u}[\Delta_u]
    -\gamma_1\mathbb{E}_{I\in\mathcal{V}_u}\|E_\phi(I)-I\|_1\\
    &-\gamma_p\mathbb{E}_{I\in\mathcal{V}_u}
    d_{\mathrm{LPIPS}}(E_\phi(I),I).
\end{split}
    \label{eq:model_selection}
\end{equation}
Here $\gamma_1,\gamma_p\geq0$ weight pixel and perceptual deviations, respectively. The fixed-hyperparameter (fixed-HP) configuration sets $\gamma_1=\gamma_p=0$ and therefore selects the checkpoint with the largest validation preference gain. Shared enhancer hyperparameter optimization (HPO) uses positive fidelity penalties and applies the full regularized criterion.
The final personalized parameters are:
\begin{equation}
    \phi_u=\arg\max_{\phi\in\mathcal{H}_u}Q_u(\phi).
    \label{eq:final_checkpoint}
\end{equation}
Positive $\gamma_1$ and $\gamma_p$ favor user-specific improvement while rejecting checkpoints that obtain it through excessive visual deviation. The fixed-HP configuration supplies a prespecified preference-gain reference, while shared enhancer HPO exposes the attainable proxy-preference--fidelity trade-off. Both configurations learn $\phi_u$ from ordinary local photographs without paired personalized retouching targets.

\subsection{Personalized On-Device Inference}

For a new photograph, deployment retains only the 0.293M-parameter enhancer
$E_{\phi_u}$, removing the personalized scorer and its feature extractors after
adaptation. In a single forward pass, the lightweight coefficient predictor
estimates image-adaptive coefficients, reconstructs and combines the compressed
shared LUT bases, and applies the resulting image-wide personalized
color-grading transform. This path requires neither scorer evaluation nor
per-image optimization, user feedback, or federated communication. Its compact
single-pass design supports mobile and real-time deployment potential.
Appendix Secs.~E.4 and H give the resource accounting and benchmark scope.

% ---- Experiments ----
\section{Experiments and Results}
\label{sec:experiments}

\subsection{Experimental Setup}
FedPAIE uses a pretrained CLUT-Net checkpoint trained on
MIT-Adobe FiveK~\cite{bychkovsky2011learning} as the common CLUT enhancer
initialization supplied by Generic Enhancement Prior Learning. FiveK contains
5,000 images and five expert retouches. Flickr-AES~\cite{ren2017personalized} contains approximately
40,000 images rated by 210 users. We reserve 37 users for open-world evaluation.
Filtering the remainder leaves 87 clients for 20 federated rounds and 10 fixed
global-validation identities. Each user follows a fixed 70/10/10/10 training,
personalization, validation, and test split. Shot counts include only support
ratings, and enhancers require validation SRCC of at least 0.10. Appendix
Sec.~E provides the remaining protocol details.

Scoring uses MSE, SRCC, and PLCC. Enhancement uses the scorer-predicted
preference gain $\Delta_u=S_{\theta_u}(\hat I)-S_{\theta_u}(I)$ as an
optimization-aligned personalization proxy. PSNR, SSIM, and
LPIPS~\cite{zhang2018unreasonable} independently measure
input preservation or similarity to the non-personalized Expert~C reference
(the FiveK evaluation ground truth). Appendix Secs.~E.5 and H give the metric
definitions and interpretation scope.

\subsection{Main Results}
We first evaluate Federated Aesthetic Preference Learning and Personalized
Scorer Calibration. Round~13 attains the peak validation SRCC of 0.5623 and is selected for
personalization. Appendix Sec.~F reports the complete global trajectory, including
loss, MSE, PLCC, and prediction spread.

\begin{table}
\centering
\footnotesize
\setlength{\tabcolsep}{3.8pt}
\resizebox{0.8\columnwidth}{!}{%
\begin{tabular}{@{}clccc@{}}
\toprule
Support & Scorer initialization & SRCC$\uparrow$ & PLCC$\uparrow$ & MSE$\downarrow$ \\
\midrule
10 & Centralized & 0.5291 & 0.5404 & 0.0532 \\
10 & Federated, fixed HP & 0.5412 & 0.5471 & 0.0589 \\
10 & Federated, per-user HPO & 0.5413 & 0.5375 & 0.0700 \\
\midrule
100 & Centralized & 0.5690 & 0.5727 & 0.0552 \\
100 & Federated, fixed HP & 0.5649 & 0.5665 & 0.0595 \\
100 & Federated, per-user HPO & 0.5625 & 0.5646 & 0.0609 \\
\bottomrule
\end{tabular}%
}
\caption{Test performance after Personalized Scorer Calibration on unseen
users. Per-user scorer HPO fits the support set and selects on disjoint
validation data.}
\label{tab:scorer_personalization}
\end{table}

Tab.~\ref{tab:scorer_personalization} shows that the federated initialization
with fixed calibration hyperparameters gives the strongest 10-shot PLCC and
nearly ties per-user HPO in SRCC, suggesting that Federated Global Preference
Modeling supplies a useful prior for sparse Personalized Scorer Calibration.
At 100 shots, additional ratings reduce sensitivity to initialization.
Fixed calibration remains competitive in scorer-only metrics. We retain
per-user scorer HPO downstream because Appendix Sec.~G shows higher
corresponding frozen-scorer outputs and input-reference PSNR across both
enhancer configurations and support regimes. Appendix Sec.~F provides user-level
variation and convergence analysis. We next evaluate Frozen-Scorer-Guided
Enhancer Adaptation. Tab.~\ref{tab:enhancement_tradeoff} contrasts the
prespecified fixed-HP open-world reference with shared enhancer HPO for
within-cohort trade-off and matched ablation analyses. Both use the same
per-user-HPO scorers and differ only in Frozen-Scorer-Guided Enhancer
Adaptation. Further details are provided in Appendix Secs.~E and F.

\begin{table}
\centering
\footnotesize
\setlength{\tabcolsep}{1.8pt}
\resizebox{\linewidth}{!}{%
\begin{tabular}{@{}l|cc|ccc|ccc@{}}
\toprule
\multirow{2}{*}{\shortstack{Support /\\Enhancer}} &
\multicolumn{2}{c|}{Scorer proxy} &
\multicolumn{3}{c|}{Flickr-AES input reference} &
\multicolumn{3}{c}{FiveK Expert~C reference} \\
\cmidrule(lr){2-3}\cmidrule(lr){4-6}\cmidrule(l){7-9}
& Score$\uparrow$ & $\Delta$$\uparrow$ &
PSNR$\uparrow$ & SSIM$\uparrow$ & LPIPS$\downarrow$ &
PSNR$\uparrow$ & SSIM$\uparrow$ & LPIPS$\downarrow$ \\
\midrule
10 / Fixed HP & $0.5039^{t}$ & $0.0247^{v}$ &
30.43 & 0.9627 & 0.0162 & 17.97 & 0.801 & 0.130 \\
10 / HPO$^{\dagger}$ & $0.5624^{t}$ & $0.0826^{t}$ &
20.51 & 0.7665 & 0.1206 & 18.33 & 0.777 & 0.147 \\
100 / Fixed HP & $0.5245^{t}$ & $0.0244^{v}$ &
31.12 & 0.9716 & 0.0132 & 18.12 & 0.804 & 0.129 \\
100 / HPO$^{\dagger}$ & $0.5864^{t}$ & $0.0858^{t}$ &
19.76 & 0.7419 & 0.1348 & 18.58 & 0.775 & 0.147 \\
\bottomrule
\end{tabular}%
}
\caption{Personalized enhancement trade-off. Input-reference metrics use
Flickr-AES, while metrics relative to Expert~C use FiveK. $^{\dagger}$ Shared
enhancer HPO selects on validation partitions from eight evaluation identities
without using test images. Superscripts $v/t$ mark validation/test proxy
quantities. Proxy and fidelity axes are distinct.}
\label{tab:enhancement_tradeoff}
\end{table}

\begin{table}[!b]
\centering
\footnotesize
\setlength{\tabcolsep}{1.6pt}
\resizebox{0.85\columnwidth}{!}{%
\begin{tabular}{@{}lccccc@{}}
\toprule
Method & Train./src. & PSNR$\uparrow$ & SSIM$\uparrow$ &
LPIPS$\downarrow$ & Params. \\
\midrule
SpliNet & C/R & 18.74 & 0.819 & -- & 0.03M \\
PieNet & C/R & 20.52 & 0.850 & -- & 28M \\
Masked Style Model. & C/R & 22.98 & 0.897 & -- & 90M \\
\midrule
Original input & --/S & 17.84 & 0.791 & 0.138 & -- \\
AdaInt + pers. scorer & C+L/S & \textbf{19.45} & 0.756 & 0.231 &
$\sim$0.6M \\
FedPAIE, 10-shot & F+L/S & 17.97 & 0.801 & 0.130 & 0.293M \\
FedPAIE, 100-shot & F+L/S & 18.12 & \textbf{0.804} &
\textbf{0.129} & 0.293M \\
\bottomrule
\end{tabular}%
}
\caption{Personalized enhancement comparison. C, F, and L denote centralized,
federated, and local training, while R and S denote literature-reported and
study-evaluated results. Bold indicates the best result among S rows.}
\label{tab:enhancement_comparison}
\end{table}

\begin{figure}[t]
\centering
\includegraphics[width=\columnwidth]{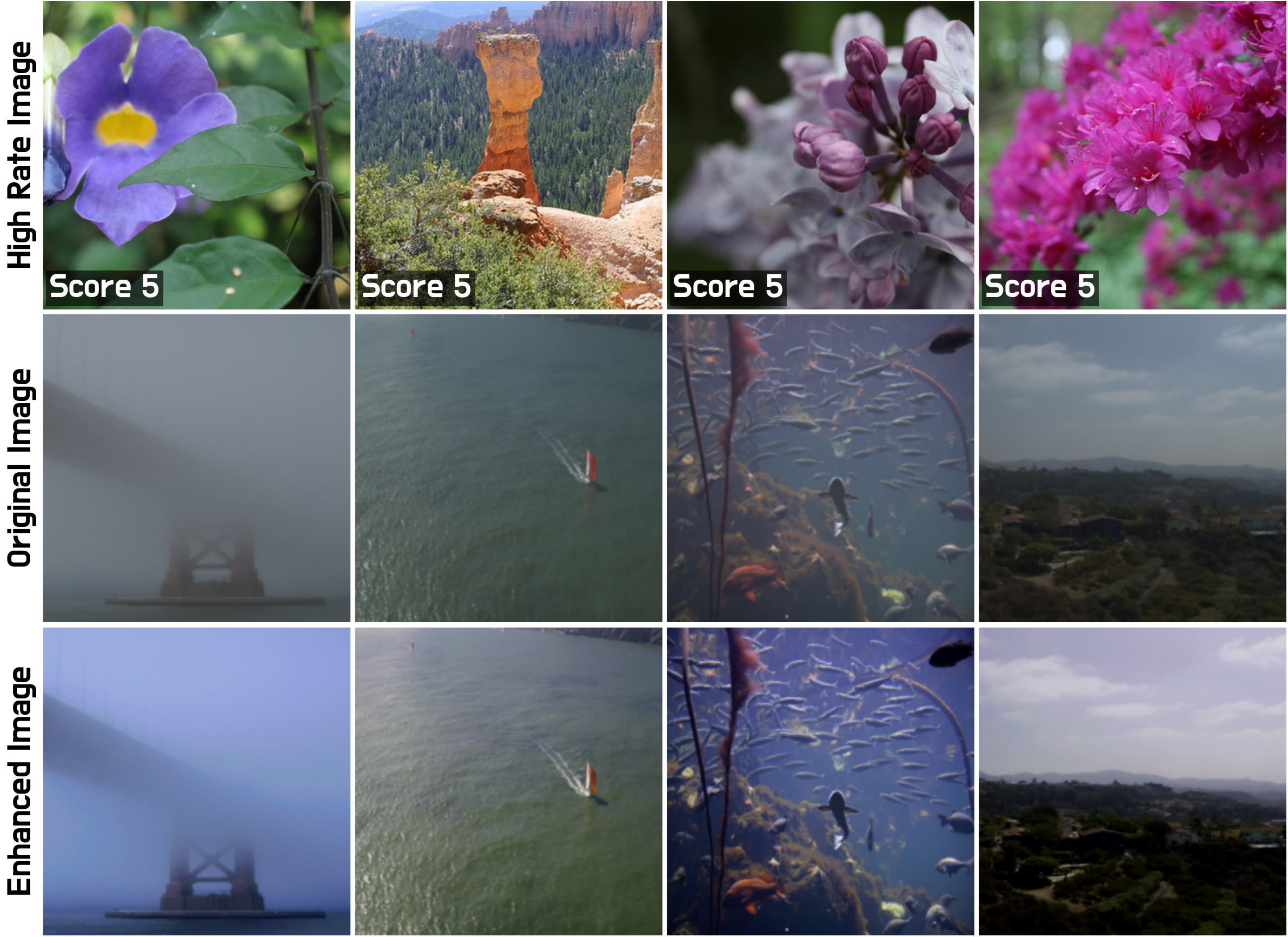}
\caption{Flickr-AES user~204 (100-shot). Rows show Flickr examples rated 5 out of 5,
FiveK inputs, and Full-objective outputs from Frozen-Scorer-Guided Enhancer
Adaptation. The Flickr examples are unpaired preference context.}
\label{fig:user204_preference}
\end{figure}

\begin{table}[t]
\centering
\footnotesize
\renewcommand{\arraystretch}{0.96}
\setlength{\tabcolsep}{1.4pt}
\resizebox{\columnwidth}{!}{%
\begin{tabular}{@{}l|ccccc|ccccc@{}}
\toprule
& \multicolumn{5}{c|}{\textit{10-shot}} &
\multicolumn{5}{c}{\textit{100-shot}} \\
\cmidrule(lr){2-6}\cmidrule(lr){7-11}
Variant & Score$\uparrow$ & $\Delta$$\uparrow$ & $P_C\!\uparrow$ &
$S_C\!\uparrow$ & $L_C\!\downarrow$ &
Score$\uparrow$ & $\Delta$$\uparrow$ & $P_C\!\uparrow$ &
$S_C\!\uparrow$ & $L_C\!\downarrow$ \\
\midrule
Original
& 0.4324 & 0.0000 & -- & -- & --
& 0.4534 & 0.0000 & -- & -- & -- \\
Generic prior
& 0.4939 & +0.0615 & 22.60 & 0.904 & 0.087
& 0.5153 & +0.0619 & 22.60 & 0.904 & 0.087 \\
Full objective
& 0.5213 & +0.0890 & 18.33 &
0.777 & 0.147
& 0.5438 & +0.0904 & 18.58 &
0.775 & 0.147 \\
$-\mathcal L_{\mathrm{pref}}$
& 0.5160 & +0.0836 & 18.46 & 0.789 & 0.139
& 0.5436 & +0.0903 & 18.45 & 0.775 & 0.145 \\
$-\mathcal L_{\mathrm{gap}}$
& 0.5295 & +0.0971 & 13.43 & 0.495 & 0.341
& 0.5513 & +0.0979 & 14.23 & 0.534 & 0.310 \\
$-$ reg. group
& 0.5173 & +0.0850 & 7.30 & 0.235 & 0.548
& 0.5269 & +0.0736 & 7.73 & 0.289 & 0.529 \\
$-$ scorer guid.
& 0.4326 & +0.0002 & 18.05 & 0.814 & 0.122
& 0.4535 & +0.0001 & 18.04 & 0.814 & 0.122 \\
\bottomrule
\end{tabular}%
}
\caption{Objective ablation for Frozen-Scorer-Guided Enhancer Adaptation.
Score is the mean frozen personalized scorer output, and $\Delta$ is its
change from Original. $P_C$, $S_C$, and $L_C$ denote PSNR, SSIM, and LPIPS
relative to Expert~C.
``$-$ reg. group'' removes $\mathcal L_1$,
$\mathcal L_{\mathrm{perc}}$, and $\mathcal L_{\mathrm{gap}}$. Generic prior
denotes the CLUT enhancer obtained through Generic
Enhancement Prior Learning. Proxy and fidelity axes jointly characterize the
preference--fidelity trade-off.}
\label{tab:objective_ablation}
\end{table}

We include SpliNet~\cite{bianco2020personalized}, PieNet~\cite{kim2020pienet},
and Masked Style Modeling~\cite{kosugi2024personalized} as literature context,
and evaluate a controlled personalized AdaInt baseline under our shared
protocol~\cite{yang2022adaint}. These rows provide literature context or
same-protocol evidence. Appendix Sec.~I distinguishes newer task settings that
are not directly comparable. The fixed-HP configuration
emphasizes input fidelity, while shared enhancer HPO explores stronger
transformations under the regularized selection criterion. Against Expert~C,
shared enhancer HPO yields higher PSNR, whereas fixed HP retains higher SSIM and
lower LPIPS, revealing complementary proxy--fidelity operating points.
Fig.~\ref{fig:qualitative_main} shows cooler, neutral, and darker
client-dependent transforms without spatial modification, as expected from
user-specific color grading. Fig.~\ref{fig:user204_preference} adds four
5/5 examples as user~204's preference context alongside 100-shot personalized
outputs. Appendix Sec.~G gives
extended analyses. Under the shared Expert~C protocol in
Tab.~\ref{tab:enhancement_comparison}, FedPAIE raises SSIM from 0.756 to
0.801/0.804 and reduces LPIPS from 0.231 to 0.130/0.129 for 10/100 shots.
AdaInt retains 1.48/1.33~dB higher PSNR, showing complementary pixel- and
perceptual-fidelity operating points. FedPAIE also improves all three metrics
over the input and uniquely combines Federated Aesthetic Preference Learning
with Frozen-Scorer-Guided Enhancer Adaptation. Its 0.293M-parameter enhancer
uses roughly half as many parameters as the study-evaluated personalized
AdaInt baseline and about $1/96$ and $1/307$ as many as PieNet and Masked Style
Modeling, respectively. Literature-sourced rows broaden
the quality--model-scale context, while the study-evaluated rows provide the
same-protocol evidence detailed in Appendix Sec.~F.

Training and Personalized On-Device Inference remain lightweight. Federated
Aesthetic Preference Learning, 10-shot Personalized Scorer
Calibration, and Frozen-Scorer-Guided Enhancer Adaptation update at most
0.787M, 0.527M, and 0.265M parameters, respectively. Static costs are
4.70~MFLOPs per scorer sample and 5.16~GFLOPs per $224\times224$ enhancer
image. Personalized On-Device Inference retains only the 0.293M-parameter
enhancer. Appendix Sec.~E.4 gives complete resource accounting, while Appendix
Sec.~G provides the image-conditioned variation and image-suitability analyses.

\subsection{Ablation Study}
\label{sec:ablation}
\begin{figure}[t]
\centering
\includegraphics[width=\columnwidth]{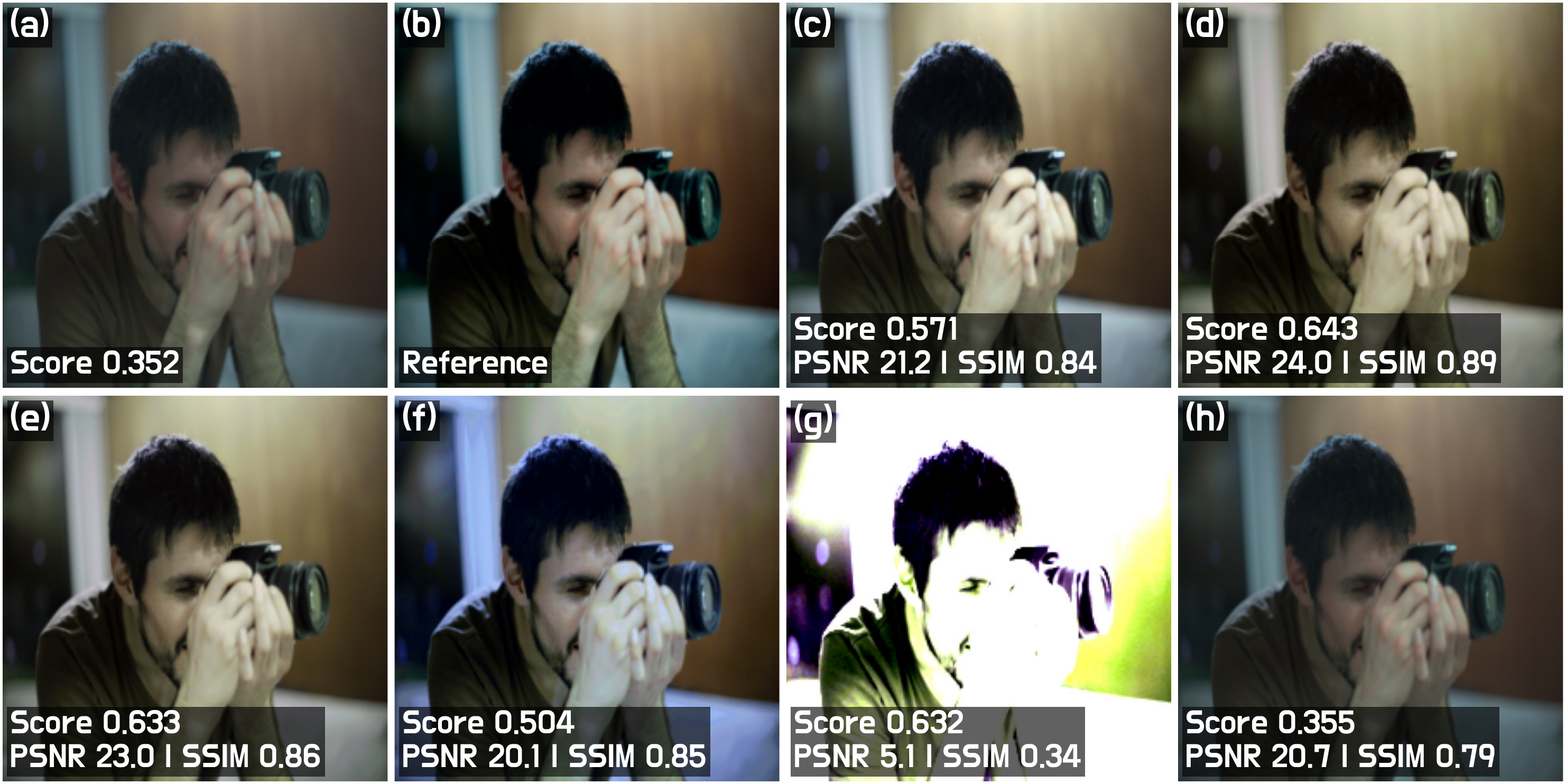}
\caption{Qualitative objective ablation for user~204 (100-shot): (a) Original,
(b) Expert~C reference, (c) Generic Enhancement Prior, (d) Full objective,
(e) w/o the preference-ranking loss $\mathcal L_{\mathrm{pref}}$, (f) w/o the
excess-gap penalty $\mathcal L_{\mathrm{gap}}$, (g) w/o the fidelity-plus-gap
regularization group $(\mathcal L_1,\mathcal L_{\mathrm{perc}},
\mathcal L_{\mathrm{gap}})$, and (h) w/o frozen-scorer guidance
$(\mathcal L_{\mathrm{pref}},\mathcal L_{\mathrm{aes}},
\mathcal L_{\mathrm{gap}})$. With matched settings, Full improves both the
frozen-scorer output and Expert~C fidelity over the Generic Enhancement Prior.}
\label{fig:qualitative_ablation}
\end{figure}

All Frozen-Scorer-Guided Enhancer Adaptation variants share the data, frozen
personalized scorers, eligibility rule, optimization protocol, and the shared
enhancer HPO configuration selected for the Full objective. Holding them fixed isolates the
active terms in Equation~\eqref{eq:personal_enhancer} without variant-specific
search. These matched runs test objective-term contributions at a common
within-cohort operating point, while fixed HP supplies the open-world reference.
Appendix Sec.~G gives the full protocol. Tab.~\ref{tab:objective_ablation} and
Fig.~\ref{fig:qualitative_ablation}
compare the Full objective with matched removal variants. Scorer guidance drives
the scorer-predicted preference gain, while $\mathcal L_{\mathrm{gap}}$ and the fidelity-plus-gap
regularization group protect fidelity. Complete results appear in Appendix
Sec.~G. Under the frozen personalized-scorer proxy, Full exceeds the Generic
Enhancement Prior for every evaluated user in both support regimes, with paired tests yielding
$p\leq3.6\times10^{-6}$.

% ---- Conclusion ----
\section{Conclusion}

We presented FedPAIE, a federated framework for personalized aesthetic image
enhancement that keeps raw photos and ratings local. Federated preference
learning initializes a scorer, which is calibrated and frozen to guide unpaired
local CLUT adaptation. The pipeline updates at most 0.787M scorer and 0.265M
enhancer parameters and retains only the 0.293M enhancer at inference.
Experiments demonstrate effective 10- and 100-shot personalization. Controlled
ablations identify scorer guidance as the preference signal and fidelity-plus-gap
regularization as the mechanism balancing preference gain with image fidelity.
FedPAIE thus provides a lightweight path from decentralized aesthetic feedback
to personalized image transformation.

% The main paper and appendix share this single bibliography after the main text.
\bibliography{references}

% Supplementary content follows the shared bibliography.
\appendix
\twocolumn[{
    \centering
    {\LARGE\bfseries Appendix\par}
    \vspace{1em}
}]
% Template-independent supplementary content.

\section{End-to-End Optimization Procedure}
\label{app:optimization}

The main paper defines the learning objectives of FedPAIE. This appendix
complements those definitions with the execution order, parameter-update
rules, and compressed-LUT construction needed to reproduce the method. We
retain the separation between two shared initializations and private
adaptation. Federated Aesthetic Preference Learning and Generic Enhancement
Prior Learning proceed independently, while On-Device Preference Adaptation
first performs Personalized Scorer Calibration and then uses the frozen
personalized scorer in Frozen-Scorer-Guided Enhancer Adaptation.

\subsection{Global Initialization}

Algorithm~\ref{alg:global} summarizes global initialization. At each
communication round, the server transmits only the current global scorer
parameters to the selected clients. A client updates the scorer using only rating
regression on its private image--rating samples and returns the updated scorer
parameters. Pairwise ordering and variance preservation are not part of this
federated objective. They are introduced only during Personalized Scorer
Calibration. The generic enhancer is trained on a separate paired retouching
corpus and therefore does not participate in the federated exchange. Client
$k$ owns $\mathcal D_k=\{(I_i,y_i)\}_{i=1}^{n_k}$, where $I_i$ is an image,
$y_i\in[0,1]$ is its normalized rating, and $n_k=|\mathcal D_k|$.
$\theta_k$ denotes the client's current scorer parameters. At round $t$, $\theta^t$
denotes the server scorer parameters before local training, $\mathcal C_t$
the participating client set, and $\theta_k^{t+1}$ the parameters returned by
client $k$ after local training.

For a local mini-batch $\mathcal B\subset\mathcal D_k$, the complete global
mean-squared-error (MSE) scorer objective is:
\begin{equation}
\begin{split}
    \mathcal L_{\mathrm{FL}}^k(\mathcal B)
    &=\mathcal L_{\mathrm{reg}}^k(\mathcal B)\\
    &=\frac{1}{|\mathcal B|}\sum_{(I_i,y_i)\in\mathcal B}
    \omega_i\bigl(S_{\theta_k}(I_i)-y_i\bigr)^2.
\end{split}
    \label{eq:app_global_regression}
\end{equation}
The default is $\omega_i=1$. Optional inverse-frequency weights only rebalance
the regression errors. Equation~\eqref{eq:app_global_regression} remains an
MSE-only objective and contains no ranking or variance term.

\begin{algorithm}[!t]
\scriptsize
\caption{Global Initialization of FedPAIE}
\label{alg:global}
\begin{algorithmic}[1]
\REQUIRE Rated sets for $K$ clients $\{\mathcal D_k\}_{k=1}^{K}$, pretrained
generic enhancer parameters $\phi^g$, rounds $T$, local epochs $E_s$, and
learning rate $\eta_s$
\ENSURE Global scorer $S_{\theta^g}$ and generic enhancer $E_{\phi^g}$
\STATE Initialize scorer parameters $\theta^0$
\FOR{$t=0,\ldots,T-1$}
    \STATE Sample participating clients $\mathcal C_t$
    \FOR{each $k\in\mathcal C_t$ in parallel}
        \STATE Set $\theta_k\leftarrow\theta^t$
        \STATE Set processed-example count $m_k\leftarrow0$
        \FOR{$e=1,\ldots,E_s$}
            \FOR{each local mini-batch $\mathcal B\subset\mathcal D_k$}
                \STATE Compute optional regression weights
                $\{\omega_i\}_{i\in\mathcal B}$ using
                Equation~\eqref{eq:app_weights}
                \STATE Evaluate $\mathcal L_{\mathrm{FL}}^k(\mathcal B)$
                using Equation~\eqref{eq:app_global_regression}
                \STATE $\theta_k\leftarrow\theta_k-\eta_s
                \nabla_{\theta_k}\mathcal L_{\mathrm{FL}}^k(\mathcal B)$
                \STATE $m_k\leftarrow m_k+|\mathcal B|$
            \ENDFOR
        \ENDFOR
        \STATE Return $\theta_k$ and the processed-example count $m_k$
    \ENDFOR
    \STATE $\displaystyle \theta^{t+1}\leftarrow
    \sum_{k\in\mathcal C_t}
    \frac{\sqrt{m_k}}{\sum_{j\in\mathcal C_t}\sqrt{m_j}}\theta_k$
\ENDFOR
\STATE $\theta^g\leftarrow\theta^T$
\RETURN $\theta^g,\phi^g$
\end{algorithmic}
\end{algorithm}

\subsection{On-Device Preference Adaptation}

For a new user, Personalized Scorer Calibration must precede
Frozen-Scorer-Guided Enhancer Adaptation. This
ordering prevents the preference target from drifting while the enhancer is
being optimized. Algorithm~\ref{alg:personalization} makes the separation
explicit. The Support-Dependent Scorer Mask is defined in
Equation~\eqref{eq:app_scorer_mask}. The enhancer mask $\mathbf M^E$ always
freezes the compressed LUT bases. Checkpoints are evaluated on a local
validation subset using $Q_u$. In the fixed-hyperparameter (fixed-HP)
configuration,
$\gamma_1=\gamma_p=0$, so $Q_u$ reduces to validation preference gain. The
shared enhancer hyperparameter optimization (HPO) configuration uses positive
fidelity penalties and therefore applies the full regularized criterion. No
server interaction is required after the two shared initializations have been
downloaded. The symbol $\odot$ denotes elementwise
multiplication in the masked updates below. For compactness, the calibration
objective in the algorithm is denoted by:
\begin{equation}
    \mathcal L_u^S
    =\lambda_{\mathrm{reg}}^u\mathcal L_{\mathrm{reg}}^u
    +\widetilde{\lambda}_{\mathrm{pair}}^u
    \mathcal L_{\mathrm{pair}}^u
    +\lambda_{\mathrm{var}}^u\mathcal L_{\mathrm{var}}^u.
    \label{eq:app_calibration}
\end{equation}
The nonnegative coefficients $\lambda_{\mathrm{reg}}^u$,
$\widetilde{\lambda}_{\mathrm{pair}}^u$, and $\lambda_{\mathrm{var}}^u$ weight
regression, pairwise ordering, and variance preservation, respectively. The
effective pairwise coefficient can be zero when the support set is too small to
provide reliable ordering supervision. Its support-regime selection and
optional collapse protection are specified below.

\begin{algorithm}
\footnotesize
\caption{On-Device Preference Adaptation for User $u$}
\label{alg:personalization}
\begin{algorithmic}[1]
\REQUIRE Shared initializations $(\theta^g,\phi^g)$, rated support set
$\mathcal D_u^s$, unpaired images $\mathcal U_u$, validation set
$\mathcal V_u$, and scorer/enhancer learning rates $\eta_u^S,\eta_u^E$
\ENSURE Personalized enhancer $E_{\phi_u}$ for Personalized On-Device Inference
\STATE Select scorer mask $\mathbf M_u^S$ from the prespecified support regime
\STATE Initialize $\theta\leftarrow\theta^g$
\FOR{each scorer-calibration step}
    \STATE Sample $\mathcal B_s\subset\mathcal D_u^s$ and construct
    $\mathcal P(\mathcal B_s)$
    \STATE Compute $\mathcal L_{\mathrm{reg}}^u$,
    $\mathcal L_{\mathrm{pair}}^u$, and $\mathcal L_{\mathrm{var}}^u$
    \STATE $\mathbf g_\theta\leftarrow\mathbf M_u^S\odot
    \nabla_{\theta}\mathcal L_u^S(\mathcal B_s)$
    \STATE $\theta\leftarrow\theta-\eta_u^S\mathbf g_\theta$
\ENDFOR
\STATE Set $\theta_u\leftarrow\theta$ and freeze $S_{\theta_u}$
\STATE Initialize $\phi\leftarrow\phi^g$ and freeze the LUT factors $\beta$
\STATE Set best validation score $Q_{\mathrm{best}}\leftarrow-\infty$
\FOR{each enhancer-adaptation epoch}
    \FOR{each mini-batch $\mathcal B_e\subset\mathcal U_u$}
        \STATE Compute $\hat I=E_\phi(I)$ and
        $\Delta_u=S_{\theta_u}(\hat I)-S_{\theta_u}(I)$
        \STATE $\mathbf g_\phi\leftarrow\mathbf M^E\odot
        \nabla_{\phi}\mathcal L_u^E(\mathcal B_e)$
        \STATE $\phi\leftarrow\phi-\eta_u^E\mathbf g_\phi$
    \ENDFOR
    \STATE Evaluate $Q_u(\phi)$ on $\mathcal V_u$
    \IF{$Q_u(\phi)>Q_{\mathrm{best}}$}
        \STATE Save $\phi_u\leftarrow\phi$ and
        $Q_{\mathrm{best}}\leftarrow Q_u(\phi)$
    \ENDIF
\ENDFOR
\STATE Discard $S_{\theta_u}$
\RETURN $E_{\phi_u}$
\end{algorithmic}
\end{algorithm}

\section{Operational Objectives and Gradient Routing}
\label{app:operational}

\subsection{Federated Global Preference Modeling: Regression and Rating Rebalancing}

Global federated training uses Equation~\eqref{eq:app_global_regression} only.
To make its optional inverse-frequency weighting unambiguous, let
$\{\mathcal R_b\}_{b=1}^{B_r}$ be a fixed partition of the normalized rating
range, let $n_{k,b}$ be the number of client-$k$ samples in bin $b$, and let
$\mathcal B_k^+=\{b:n_{k,b}>0\}$. Here $B_r$ is the number of bins, and $b(i)$
denotes the bin containing rating
$y_i$. We first normalize inverse-frequency class
weights by their mean over nonempty bins and then clip them for stability:
\begin{equation}
\begin{split}
    r_{k,b}&=\frac{n_k}{n_{k,b}},\qquad
    \bar r_k=\frac{1}{|\mathcal B_k^+|}
    \sum_{b\in\mathcal B_k^+}r_{k,b},\\
    \omega_i&=\operatorname{clip}\!\left(
    \frac{r_{k,b(i)}}{\bar r_k},
    \omega_{\min},\omega_{\max}\right).
\end{split}
    \label{eq:app_weights}
\end{equation}
Weighting is activated only when the largest rating-bin proportion exceeds a
prespecified imbalance threshold $\xi$. Here $\xi$ is the activation threshold
and $\omega_{\min},\omega_{\max}$ are the lower and upper clipping bounds.
Otherwise, all $\omega_i$ are set to one. Empty bins receive no samples and
therefore do not contribute to the
mini-batch loss. The bin boundaries, $\xi$, $\omega_{\min}$, and
$\omega_{\max}$ are implementation hyperparameters and are reported with the
experimental settings. Crucially, activating these weights changes only how
the squared errors are averaged. It does not add a second training signal.

\subsection{Personalized Pairwise and Variance Objectives}

Pair construction is used only after global training, when the scorer is
calibrated to a new user. For a local support mini-batch
$\mathcal B_s=\{(I_i,y_i)\}_{i=1}^{B_s}$ of size $B_s$, let $\delta>0$ be the
minimum normalized-rating separation. We use each unordered pair once and
exclude pairs whose ratings are too close to provide a reliable direction:
\begin{equation}
    \mathcal P(\mathcal B_s)
    =\{(i,j):1\leq i<j\leq B_s,\ |y_i-y_j|>\delta\}.
    \label{eq:app_pairs}
\end{equation}
With $r_{ij}=\operatorname{sign}(y_i-y_j)$, the pairwise term is:
\begin{equation}
    \mathcal L_{\mathrm{pair}}^u
    =-\frac{1}{|\mathcal P(\mathcal B_s)|}
    \sum_{(i,j)\in\mathcal P(\mathcal B_s)}
    \log\sigma\!\left(r_{ij}(\hat y_i-\hat y_j)\right).
    \label{eq:app_pair_loss}
\end{equation}
Here $\sigma$ is the sigmoid function.
We set $\mathcal L_{\mathrm{pair}}^u=0$ when
$\mathcal P(\mathcal B_s)=\varnothing$. Let $\mathbf y$ and
$\hat{\mathbf y}$ collect the target and predicted ratings in $\mathcal B_s$.
The variance term is:
\begin{equation}
    \mathcal L_{\mathrm{var}}^u
    =\left[\rho\,\operatorname{Std}(\mathbf y)
    -\operatorname{Std}(\hat{\mathbf y})\right]_+,
    \label{eq:app_variance_loss}
\end{equation}
where $[z]_+=\max(z,0)$. It penalizes predictions whose dispersion falls below
a fraction $\rho\in(0,1]$ of the observed rating dispersion. The operational implementation uses
$\delta=0.1$ and $\rho=0.7$ for ratings normalized to $[0,1]$.

To reduce the influence of unreliable ordering gradients when the scorer is
close to a constant predictor, the effective pairwise coefficient can be
attenuated according to:
\begin{equation}
    \widetilde{\lambda}_{\mathrm{pair}}^u=
    \begin{cases}
    \kappa\lambda_{\mathrm{pair}}^u,
    & \operatorname{Std}(\hat{\mathbf y})<\epsilon_c,\\
    \lambda_{\mathrm{pair}}^u, & \text{otherwise},
    \end{cases}
    \label{eq:app_effective_pair_weight}
\end{equation}
where $\kappa$ is the attenuation factor and $\epsilon_c$ the collapse
threshold. The implementation uses $\kappa=0.5$ and $\epsilon_c=0.01$. If the
support-regime coefficient $\lambda_{\mathrm{pair}}^u$ is zero, the pairwise
term remains disabled regardless of this rule. Thus, an empty-pair mini-batch
still contributes through regression and variance preservation.

\subsection{Masked Parameter Updates}

Write the scorer parameters as
$\theta=(\theta_c,\theta_s,\theta_f,\tau)$, corresponding to the color
projection, semantic projection, fusion head, and temperature. The pretrained
MobileNetV3 semantic extractor and the deterministic color-statistics extractor
are fixed and are not included in this trainable tuple. Let
$N_u=|\mathcal D_u^s|$. The implementation uses $N_0=20$ to separate the
small- and larger-support regimes. The personalization mask is:
\begin{equation}
    \mathbf M_u^S=
    \begin{cases}
    \mathbf 1_{\theta_f}+\mathbf 1_{\tau}, & N_u\leq N_0,\\
    \mathbf 1_{\theta_c}+\mathbf 1_{\theta_s}
    +\mathbf 1_{\theta_f}+\mathbf 1_{\tau}, & N_u>N_0,
    \end{cases}
    \label{eq:app_scorer_mask}
\end{equation}
where $\mathbf 1_{a}$ selects the coordinates of parameter block $a$ and
$\mathbf 0_a$ is zero on that block. All unselected coordinates are zero.
Consequently, the 10-shot configuration
updates only the fusion MLP and temperature, whereas the 100-shot configuration
also updates both scorer projections. The scorer update at calibration step
$\ell$ is therefore:
\begin{equation}
    \theta^{\ell+1}
    =\theta^{\ell}-\eta_u^S\mathbf M_u^S\odot
    \nabla_{\theta}\mathcal L_u^S,
    \label{eq:app_scorer_update}
\end{equation}
which guarantees that the feature extractors and every masked scorer block
retain their global values exactly.

Similarly, decompose the enhancer as $\phi=(\psi,\beta)$, where $\psi$
contains the CNN backbone and coefficient head of the lightweight coefficient
predictor and $\beta$ contains all factorized LUT parameters. Enhancer
adaptation uses:
\begin{equation}
\begin{split}
    \mathbf M^E&=(\mathbf 1_{\psi},\mathbf 0_{\beta}),
    \qquad \beta^{\ell+1}=\beta^g,\\
    \psi^{\ell+1}&=\psi^\ell-\eta_u^E
    \nabla_{\psi}\mathcal L_u^E.
\end{split}
    \label{eq:app_enhancer_mask}
\end{equation}
Although $S_{\theta_u}$ is frozen, it remains in the differentiable path. For
the preference term $\ell_{\mathrm{pref}}=-\log\sigma(\Delta_u)$, its gradient
with respect to the trainable enhancer block is:
\begin{equation}
\begin{split}
    \nabla_{\psi}\ell_{\mathrm{pref}}
    ={}&(\sigma(\Delta_u)-1)
    \mathbf J_{E,\psi}(I)^{\top}\\
    &\cdot\nabla_{\hat I}S_{\theta_u}(\hat I),
\end{split}
    \label{eq:app_gradient_flow}
\end{equation}
where $\mathbf J_{E,\psi}$ is the enhancer Jacobian. Equation
\eqref{eq:app_gradient_flow} clarifies that the scorer supplies image-space
gradients without receiving a parameter update.

\section{Compressed LUT Parameterization}
\label{app:clut}

Let each residual LUT basis contain three color channels on a grid of
resolution $d$. In the fully factorized form, the channel-$c$ tensor of basis
$q$ is reconstructed from two shared factors and a basis-specific core:
\begin{equation}
    \mathbf B_{q,c}(\beta)
    =\operatorname{reshape}_{d\times d\times d}
    \left(\mathbf A\mathbf C_{q,c}\mathbf D\right),
    \label{eq:app_lut_factorization}
\end{equation}
where $r_s$ and $r_w$ are factorization ranks,
$\mathbf A\in\mathbb R^{d\times r_s}$,
$\mathbf C_{q,c}\in\mathbb R^{r_s\times r_w}$, and
$\mathbf D\in\mathbb R^{r_w\times d^2}$. The collection
$\beta=\{\mathbf A,\mathbf D,\mathbf C_{q,c}\}_{q,c}$ parameterizes all
bases. Here $M$ is the number of LUT bases. For an image $I$, the lightweight
coefficient predictor, comprising a CNN backbone and coefficient head, produces
$\mathbf w_\psi(I)\in\mathbb R^M$, giving:
\begin{equation}
\begin{split}
    \mathbf L_{\psi,\beta}(I)
    & =\sum_{q=1}^{M}w_{\psi,q}(I)\mathbf B_q(\beta),\\
    \widetilde I
    & =I+\mathcal T(I,\mathbf L_{\psi,\beta}(I)),\\
    \hat I&=\operatorname{clip}\!\left(\widetilde I,0,1\right).
\end{split}
    \label{eq:app_lut_application}
\end{equation}
$\mathcal T$ denotes trilinear interpolation of the fused LUT at the input RGB
values. The CLUT-Net forward path returns $\widetilde I$. Its Paired Training
Objective is evaluated on this unclipped output. The clipping step is
applied externally during Frozen-Scorer-Guided Enhancer Adaptation and
Personalized On-Device Inference. The pretrained checkpoint supplies both
$\psi^g$ and $\beta^g$. During personalization, fixing $\beta^g$ preserves the
learned space of plausible color transforms, while updating $\psi$ changes how
an image is mapped to a mixture of those transforms. This restriction is the
architectural counterpart to the fidelity and excess-gap penalties in the
Personalized Enhancement Objective.

\section{Component Lifecycle and Privacy Boundary}
\label{app:lifecycle}

Tab.~\ref{tab:lifecycle} collects the trainable/frozen status that is spread
across the main method. Its local column reports Personalized Scorer Calibration
and Frozen-Scorer-Guided Enhancer Adaptation in that order. The conditional update follows the
support-size rule in Equation~\eqref{eq:app_scorer_mask}.

\begin{table}
\centering
\scriptsize
\setlength{\tabcolsep}{3pt}
\begin{tabular}{@{}lccc@{}}
\toprule
Component & Global & Local $S/E$ & Deploy \\
\midrule
Color statistics $\Phi_c$ & F & F/UF & D \\
Semantic extractor $\Phi_s$ & F & F/UF & D \\
Color projection $P_c$ & U-FL & U$^*$/UF & D \\
Semantic projection $P_s$ & U-FL & U$^*$/UF & D \\
Fusion head $G_{\theta_f}$ & U-FL & U/UF & D \\
Temperature $\tau$ & U-FL & U/UF & D \\
Enhancer predictor $\psi$ & U-P & --/U & R \\
Compressed LUT factors $\beta$ & U-P & --/F & R \\
\bottomrule
\end{tabular}
\caption{Lifecycle of FedPAIE components. The local column reports Personalized
Scorer Calibration/Frozen-Scorer-Guided Enhancer Adaptation. U, F, UF, D, and R denote updated, frozen,
used but frozen, discarded, and retained. FL and P denote federated and paired
global learning. \textsuperscript{*}Updated only in the larger-support regime.}
\label{tab:lifecycle}
\end{table}

For clarity, the server-visible state at communication round $t$ is limited to:
\begin{equation}
    \mathcal V_{\mathrm{server}}^t
    =\left(\theta^t,
    \{(\theta_k^{t+1},m_k):k\in\mathcal C_t\}\right).
    \label{eq:app_server_view}
\end{equation}
The local images, ratings, constructed preference pairs, and personalized
models are not uploaded. The generic enhancer is initialized from a separate
generic paired corpus. After downloading $(\theta^g,\phi^g)$, a new user's
support, adaptation, validation, and inference stages require no further
communication. This protocol enforces raw-data locality by communicating only
scorer parameters and aggregation counts while keeping every user-specific
asset and adaptation stage on device. The same communication boundary is
compatible with secure aggregation and differential privacy when additional
deployment protections are required.

\section{Experimental Protocol and Reproducibility}
\label{app:protocol}

\subsection{Datasets and Preprocessing}

MIT-Adobe FiveK~\cite{bychkovsky2011learning} contains 5,000 original
photographs and a retouched version from each of five experts. The preprocessing
pipeline converts inputs to RGB, transforms the supplied ProPhoto RGB images to
sRGB, rejects corrupted files, and resizes images while preserving orientation
to either $720\times480$ or $480\times720$. For Generic Enhancement Prior
Learning, FedPAIE loads a pretrained CLUT-Net checkpoint trained on
paired FiveK retouching data. All personalized enhancer conditions start from this same
checkpoint, so the source initialization is controlled across comparisons.

Flickr-AES~\cite{ren2017personalized} contains approximately 40,000 images
rated by 210 users on a 1--5 scale. Ratings are normalized to $[0,1]$. We remove
records with missing or invalid fields, duplicate user--image records, unreadable
or corrupted images, and samples that cannot be converted consistently to RGB.
The remaining data are organized by numeric client identifier. This dataset is
unpaired: ratings support preference modeling, but there is no user-specific
retouched target for enhancer personalization.

\subsection{Open-World Split and Cohort Accounting}

Users, rather than images alone, are separated for the global open-world
protocol. Of the 210 users, 173 are candidate training users and 37 are held out
from federated optimization. Each user's records are further divided into 70\%
training, 10\% personalization, 10\% validation, and 10\% test subsets with a
fixed seed. The training side is filtered for at least 100 training samples and
usable rating diversity, leaving 87 clients. All eligible training clients join
each of the 20 communication rounds. Global validation is monitored on the
held-out validation partitions of 10 fixed eligible training identities. The
experiment does not subsample a new client subset from round to round.

Tab.~\ref{tab:app_cohorts} separates the cohorts used by different analyses.
Here and below, HP denotes hyperparameters and HPO denotes hyperparameter
optimization.
The shot count refers only to rated support images used for gradient-based
scorer adaptation. Validation data are used for early stopping or model
selection, and the test split is disjoint from both. A personalized scorer is
eligible to guide the enhancer only if its validation Spearman rank correlation
(SRCC) is at least 0.10, ensuring a uniform quality threshold for enhancer
guidance. The resulting cohort counts are reported in
Tab.~\ref{tab:app_cohorts}, and each analysis keeps its eligible cohort fixed
across all compared configurations.

\begin{table}
\centering
\scriptsize
\setlength{\tabcolsep}{4pt}
\begin{tabular}{@{}lr@{}}
\toprule
Cohort or analysis & Users/models \\
\midrule
Flickr-AES users & 210 \\
Candidate FL / unseen evaluation users & 173 / 37 \\
Eligible FL clients per round & 87 \\
Global-validation monitoring identities & 10 \\
10-shot enhancers: completed / skipped & 36 / 1 \\
100-shot enhancers: completed / skipped & 37 / 0 \\
Image-suitability analysis & 33 \\
\bottomrule
\end{tabular}
\caption{Cohort accounting after the stated eligibility controls. Each
analysis uses a fixed matched cohort for all reported comparisons.}
\label{tab:app_cohorts}
\end{table}

\subsection{Training and Model Selection}
Randomness is controlled by seeding Python's \texttt{random}, NumPy, the
PyTorch CPU generator, and all CUDA generators. Dataset and cohort splitting,
as well as the federated and centralized scorer runs, use seed 42. Per-user
scorer calibration uses seed $42+u$ for numeric client identifier $u$.
Fixed-HP enhancer adaptation, shared enhancer HPO, and enhancement evaluation
use seed 60. Unless otherwise stated, each internally trained model
configuration is trained once under these fixed seeds. Personalized aggregate
rows contain one trained model per eligible user. The HPO trial counts reported
below are search trials rather than repeated training seeds.
\par\noindent\textbf{Global Models.}
The same pretrained CLUT-Net checkpoint is used unchanged as $\phi^g$ for
all downstream experiments. The associated Paired Training Objective uses
the unclipped output and combines $\mathcal L_1$ with
$0.1\mathcal L_{\mathrm{perc}}$, where LPIPS denotes learned perceptual image
patch similarity. The global scorer is trained for
20 federated rounds with all eligible clients participating. Each client uses
adaptive local epochs. The returned aggregation count $m_k$ is therefore the
number of examples actually processed across its local optimization steps, as
made explicit in Algorithm~\ref{alg:global}. Square-root weighting is applied
to $m_k$. Round~13 is selected by the highest global validation Spearman rank
correlation (SRCC).
\par\noindent\textbf{Personalized Scorer Calibration.}
For each unseen user, a balanced support set of 10 or 100 local ratings is used
for adaptation, and a separate validation split selects the checkpoint.
Per-user scorer HPO uses 20 trials per user to maximize
validation SRCC. MSE and Pearson linear correlation (PLCC) are reported
metrics rather than components
of the search objective. The search covers learning rate
$[10^{-5},5\!\times\!10^{-4}]$ on a log scale, pairwise weight $[0,0.30]$,
weight decay $[10^{-7},10^{-3}]$, variance weight $[0,0.05]$, and gradient
clipping $[0.5,2.0]$. The regression coefficient is
$\max(0.60,1-\lambda_{\mathrm{pair}})$. Epoch ranges are 20--40 for 10-shot and
30--80 for 100-shot, with early stopping. The final metrics are evaluated on
the disjoint user test split.
\par\noindent\textbf{Frozen-Scorer-Guided Enhancer Adaptation.}
All four reported enhancer conditions use the corresponding personalized
scorers obtained by per-user scorer HPO. The fixed-HP and shared enhancer HPO
labels therefore refer only to
Frozen-Scorer-Guided Enhancer Adaptation, not to different Personalized Scorer
Calibration strategies.
The fixed-HP gain-selection configuration operates at $224\times224$ with
batch size 4, Adam learning rate $3\times10^{-4}$, and 40 epochs. It uses
$\lambda_{\mathrm{aes}}=0.5$, $\lambda_1=0.1$,
$\lambda_{\mathrm{perc}}=0.05$, $\lambda_{\mathrm{gap}}=3.0$, excess-gap tolerance
$\mu=0.05$, and gradient clipping $1.0$. The preference coefficient
$\lambda_{\mathrm{pref}}$ is chosen from
$\{0.01,0.03,0.05,0.09\}$ according to validation-SRCC intervals
$[0.10,0.20)$, $[0.20,0.30)$, $[0.30,0.40)$, and $[0.40,1]$, respectively.
Scorers below 0.10 are skipped. Checkpoints maximize validation preference
gain, corresponding to $\gamma_1=\gamma_p=0$ in the main-paper selection
criterion.

For shared-HPO regularized selection, 20 enhancer HPO trials are evaluated on the validation partitions of
eight users. Tab.~\ref{tab:app_enhancer_hparams} gives the selected settings,
which are then used for 40-epoch personalization. Checkpoints maximize the
regularized validation criterion from the main paper using positive fidelity
penalties. The eight validation identities are drawn from the final 37-user
cohort, while all test images remain held out. The selected configuration is
then fixed across the cohort for the within-cohort trade-off and matched
objective analyses. The prespecified fixed-HP setting provides the
user-disjoint open-world reference.

\begin{table}
\centering
\scriptsize
\setlength{\tabcolsep}{8pt}
\begin{tabular}{@{}lcc@{}}
\toprule
Hyperparameter & 10-shot & 100-shot \\
\midrule
LR & $5.4485\!\times\!10^{-4}$ & $5.3948\!\times\!10^{-4}$ \\
$\lambda_{\mathrm{pref}}$ & 0.0411 & 0.0402 \\
$\lambda_{\mathrm{aes}}$ & 0.5996 & 0.5848 \\
$\lambda_1$ & 0.1007 & 0.1271 \\
$\lambda_{\mathrm{perc}}$ & 0.0543 & 0.0543 \\
$\lambda_{\mathrm{gap}}$ & 0.5107 & 0.7642 \\
$\mu$ & 0.1048 & 0.2051 \\
Gradient clip & 1.9631 & 1.6176 \\
\bottomrule
\end{tabular}
\caption{Selected shared enhancer HPO configurations. Both use 40 final
epochs. The value $\mu$ is the excess-gap tolerance. The search used 20 trials and
selected trial 12 in both support regimes.}
\label{tab:app_enhancer_hparams}
\end{table}

Training, fine-tuning, and evaluation were conducted on a Windows~11
workstation equipped with an Intel Core Ultra~9 285K CPU, an NVIDIA RTX~5090
GPU, approximately 64~GB of system memory, and CUDA~12.8. The experimental
environment uses PyTorch~2.2.2 and
Torchvision~0.17.2. Preprocessing also used Apple M2 and M3 devices. The
deployed enhancer contains 0.293M parameters, with a measured forward time of
4.62~ms per $224\times224$ image.

\subsection{Resource Accounting}
\label{app:resource_accounting}

We derive the following resource counts from the implemented modules and update
masks. Parameter totals are exact. Arithmetic counts are static estimates that
use one multiply--accumulate (MAC) as two floating-point operations. They
exclude data loading, normalization, nonlinear activations, elementwise
HSV/Lab operations, loss reductions, optimizer bookkeeping, and validation.
Memory is reported in MiB and assumes FP32. It is an analytical model-state
lower bound rather than measured peak device memory because activations, CUDA
workspaces, dataloader buffers, and retained checkpoint copies depend on the
runtime.

\begin{table}
\centering
\scriptsize
\setlength{\tabcolsep}{2.6pt}
\begin{tabular}{@{}lrrr@{}}
\toprule
Stage & Resident & Updated & FLOPs/sample \\
\midrule
\shortstack[l]{Federated Aesthetic\\Preference Learning} & 0.787M & 0.787M & 4.70M \\
\shortstack[l]{Personalized Scorer Calibration\\(10-shot)} & 0.787M & 0.527M & 3.66M \\
\shortstack[l]{Personalized Scorer Calibration\\(100-shot)} & 0.787M & 0.787M & 4.70M \\
\shortstack[l]{Frozen-Scorer-Guided\\Enhancer Adaptation} & 6.523M & 0.265M & 5.16G \\
\shortstack[l]{Personalized On-Device\\Inference} & 0.293M & -- & -- \\
\bottomrule
\end{tabular}
\caption{Complete stage-level resource summary.
Resident counts all required weights, whereas Updated counts
gradient-updated weights. Scorer rows assume cached descriptors. The
Frozen-Scorer-Guided Enhancer Adaptation row includes frozen supervision
networks.}
\label{tab:app_resource_profile}
\end{table}

\begin{table}[!t]
\centering
\scriptsize
\setlength{\tabcolsep}{4.4pt}
\begin{tabular}{@{}lr@{}}
\toprule
Model block & Parameters \\
\midrule
Scorer color projection & 13,824 \\
Scorer semantic projection & 246,528 \\
Scorer fusion MLP & 526,849 \\
Scorer temperature & 1 \\
\shortstack[l]{\textbf{Lightweight Dual-Cue}\\\textbf{Aesthetic Scorer total}} & \textbf{787,202} \\
\midrule
CLUT CNN backbone & 245,504 \\
CLUT coefficient head & 19,092 \\
Compressed LUT bases & 27,945 \\
\textbf{CLUT-Net total} & \textbf{292,541} \\
\bottomrule
\end{tabular}
\caption{Exact parameter decomposition of the two trainable FedPAIE models.}
\label{tab:app_parameter_blocks}
\end{table}

The 10-shot scorer mask updates the fusion MLP and temperature, totaling
526,850 parameters. The 100-shot mask updates all 787,202 scorer parameters.
Frozen-Scorer-Guided Enhancer Adaptation freezes the 27,945 compressed-basis
parameters and updates the
245,504-parameter backbone and 19,092-parameter coefficient head, totaling
264,596 parameters.

\begin{table}[!t]
\centering
\scriptsize
\setlength{\tabcolsep}{2.8pt}
\begin{tabular}{@{}lcc@{}}
\toprule
Stage & Params. R/U & MiB W/T \\
\midrule
\shortstack[l]{Federated Aesthetic\\Preference Learning} & 787,202 / 787,202 & 3.003 / 12.012 \\
\shortstack[l]{Personalized Scorer Calibration\\(10-shot)} & 787,202 / 526,850 & 3.003 / 9.035 \\
\shortstack[l]{Personalized Scorer Calibration\\(100-shot)} & 787,202 / 787,202 & 3.003 / 12.012 \\
\shortstack[l]{Frozen-Scorer-Guided\\Enhancer Adaptation} & 6,522,543 / 264,596 & 24.882 / 27.910 \\
\shortstack[l]{Personalized On-Device\\Inference} & 292,541 / -- & 1.116 / 1.116 \\
\bottomrule
\end{tabular}
\caption{Parameter-state memory across the three optimization stages and
deployment. R/U denotes resident/updated parameters, and W/T denotes FP32
weights/total training state. The latter includes gradients and two optimizer
moment tensors for updated parameters. Activations and workspaces are excluded.}
\label{tab:app_stage_memory}
\end{table}

The complete Frozen-Scorer-Guided Enhancer Adaptation stack in
Tab.~\ref{tab:app_stage_memory} consists of the 0.293M-parameter CLUT-Net,
0.787M-parameter personalized scorer, 2.972M-parameter MobileNetV3 feature
trunk, and 2.471M-parameter AlexNet-LPIPS network. The latter three modules and
the compressed LUT bases are frozen. Only the 0.265M-parameter CLUT coefficient
predictor is optimized. Representative serialized files occupy 3.092~MiB for
the global scorer, 3.012~MiB for a personalized scorer, 1.129~MiB for the
generic CLUT checkpoint, and 1.125~MiB for a personalized CLUT checkpoint.
Small differences from raw FP32 weight size arise from serialization metadata.

\begin{table}
\centering
\scriptsize
\setlength{\tabcolsep}{2.5pt}
\begin{tabular}{@{}p{0.34\columnwidth}cp{0.38\columnwidth}@{}}
\toprule
Operation & MACs / FLOPs & Scope \\
\midrule
Scorer forward & 0.783M / 1.565M & All projections and fusion MLP \\
Federated Aesthetic Preference Learning & 2.348M / 4.696M & Forward and backward through all scorer blocks \\
Personalized Scorer Calibration (10-shot) & 1.832M / 3.663M & Frozen projections and trainable fusion MLP \\
\midrule
CLUT image-transform path & 0.179G / 0.358G & Forward and gradients to the coefficient predictor \\
Scorer-guidance path & 0.431G / 0.862G & MobileNetV3 and scorer gradient to the enhanced image \\
LPIPS path & 1.968G / 3.936G & Two feature passes and enhanced-branch backward \\
\textbf{Frozen-Scorer-Guided Enhancer Adaptation} & \textbf{2.578G / 5.156G} & One $224\times224$ training image \\
\bottomrule
\end{tabular}
\caption{Static core-arithmetic accounting. Batch size changes parallelism but
not the per-sample values. The estimate excludes the operations listed in the
opening paragraph of this subsection.}
\label{tab:app_stage_compute}
\end{table}
\par\noindent\textbf{Federated Aesthetic Preference Learning.}
The scorer optimizer consumes cached 24-D HSV/Lab descriptors and 960-D
MobileNetV3 embeddings. Each FP32 descriptor pair occupies approximately
3.84~KiB per image. If the semantic embedding is not already cached, its
one-time local extraction activates a 2.972M-parameter MobileNetV3 trunk and
requires approximately 0.215~GMAC per image. The resulting embedding is reused
across every local epoch and communication round.

For client $k$ with $N_k$ cached training samples, $E_k$ local epochs, and
$T=20$ communication rounds, the optimization-only computation is:
\begin{equation}
\mathcal C_k^{\mathrm{FL}}
\approx T E_k N_k (4.696~\mathrm{MFLOPs}).
\label{eq:app_fl_compute}
\end{equation}
Eligible clients with $100\leq N_k<500$ use two local epochs, while larger
clients use one. For example, $N_k=100$ and $E_k=2$ require approximately
0.94~GFLOPs per round and 18.78~GFLOPs over 20 rounds. The FP32 scorer state is
3.149~MB, or 3.003~MiB, per download or upload. Twenty rounds therefore
transfer approximately 62.98~MB in each direction and 125.95~MB bidirectionally
per participating client.
\par\noindent\textbf{Personalized Scorer Calibration.}
For a support set of $S_u$ samples and $E_u$ calibration epochs, the
optimization-only computation is:
\begin{equation}
\begin{aligned}
\mathcal C_u^{\mathrm{cal}} &\approx S_u E_u c_u,\\
c_u &=
\begin{cases}
3.663~\mathrm{MFLOPs}, & S_u=10,\\
4.696~\mathrm{MFLOPs}, & S_u=100.
\end{cases}
\end{aligned}
\label{eq:app_calibration_compute}
\end{equation}
The 20--40 epoch range for 10-shot calibration corresponds to
0.73--1.47~GFLOPs for one final fit. The 30--80 epoch range for 100-shot
calibration corresponds to 14.09--37.56~GFLOPs. The reported per-user scorer
HPO path runs 20 sequential trials followed by one final refit. Without early stopping, its
optimization-only envelope is 15.39--30.77~GFLOPs for 10-shot and
295.82--788.85~GFLOPs for 100-shot. Early stopping reduces the actual total.
These envelopes exclude repeated validation inference. Sequential trials
increase total arithmetic but do not multiply the peak model state in
Tab.~\ref{tab:app_stage_memory}.
\par\noindent\textbf{Frozen-Scorer-Guided Enhancer Adaptation.}
The original-image scorer outputs are cached, but features of the changing
enhanced image cannot be cached. We therefore count gradients through the
frozen MobileNetV3 trunk and personalized scorer. LPIPS similarly performs
feature extraction for the original and enhanced images, with backward
propagation through the enhanced branch. For $N_u$ unpaired local images and
the reported 40-epoch schedule, the optimization-only computation is:
\begin{equation}
\mathcal C_u^{E}
\approx 40 N_u (5.156~\mathrm{GFLOPs}).
\label{eq:app_enhancer_compute}
\end{equation}
Local sets of 25, 50, and 100 images therefore require approximately 5.16,
10.31, and 20.62~TFLOPs for one final adaptation. The reported shared
enhancer HPO uses 20 trials, 10 epochs per trial, and at most eight
representative users. Its corresponding optimization-only budget is
$20\times10\times\sum_u N_u\times5.156$~GFLOPs and is separate from the final
40-epoch adaptations. These static counts describe arithmetic rather than
wall-clock latency. They show that the trainable state remains compact even
when all frozen supervision networks are conservatively included.

\subsection{Metric and Reference Conventions}

For target ratings $\mathbf y$ and predictions $\hat{\mathbf y}$, scorer
evaluation uses:
\begin{equation}
\rho_{\mathrm S}=\operatorname{corr}
\bigl(\operatorname{rank}(\mathbf y),
\operatorname{rank}(\hat{\mathbf y})\bigr),\qquad
r_{\mathrm P}=\operatorname{corr}(\mathbf y,\hat{\mathbf y}),
\label{eq:app_correlations}
\end{equation}
where $\operatorname{corr}$ denotes Pearson correlation. Here,
$\rho_{\mathrm S}$ and $r_{\mathrm P}$ are SRCC and PLCC, respectively. MSE is also reported. For
user $u$, the enhancement proxy is:
\begin{equation}
\Delta_u(I)=S_{\theta_u}(E_{\phi_u}(I))-S_{\theta_u}(I).
\label{eq:app_score_gap}
\end{equation}
A positive value means that the same frozen personalized scorer used for guidance assigns a
higher score to the enhanced image. This optimization-aligned personalization
proxy is complemented by reference-based fidelity metrics and paired
user-level tests.

The image-reference metrics are peak signal-to-noise ratio (PSNR), structural
similarity (SSIM), and LPIPS.

\begin{table}
\centering
\scriptsize
\setlength{\tabcolsep}{3.5pt}
\begin{tabular}{@{}p{0.30\columnwidth}p{0.62\columnwidth}@{}}
\toprule
Quantity & Interpretation \\
\midrule
SRCC / PLCC / MSE & Agreement between predicted and observed user ratings. \\
Predicted score / scorer-predicted preference gain $\Delta$ & Personalized scorer proxy. Higher is better under that scorer. \\
PSNR/SSIM/LPIPS vs. input & Content preservation and magnitude of the applied transformation. \\
PSNR/SSIM/LPIPS vs. Expert~C & Similarity to a standardized professional retouching reference. \\
Inter-client output difference & Personalization diversity among user-specific outputs for a shared input. \\
\bottomrule
\end{tabular}
\caption{Evaluation quantities and the claims they support.}
\label{tab:app_metric_scope}
\end{table}

Unless otherwise stated, reported personalized results are means over the
valid client cohort. PSNR and SSIM are better when larger, whereas LPIPS is
better when smaller. Their meaning depends on the explicitly named reference.

\section{Extended Quantitative Results}
\label{app:extended_results}

\subsection{Federated Aesthetic Preference Learning and Personalized Scorer Calibration}

Tab.~\ref{tab:app_global_scorer} summarizes the global trajectory. Validation
MSE stabilizes after approximately 10--12 rounds. The small late-round
oscillations are consistent with non-IID data and stochastic local minibatches.
Because all eligible clients participate in every round, the trajectory
reflects optimization under the full eligible federated cohort.

\begin{table}
\centering
\scriptsize
\setlength{\tabcolsep}{6pt}
\begin{tabular}{@{}lcc@{}}
\toprule
Metric & Initial & Reported final/peak \\
\midrule
Training loss & 0.0694 & 0.0401 (round 20) \\
Validation MSE & 0.0808 & 0.0603 (round 18 minimum) \\
Validation SRCC & 0.4650 & 0.5623 (round 13 peak) \\
Validation PLCC & 0.4764 & 0.5854 (round 20 maximum) \\
Prediction standard deviation & 0.0861 & 0.1769 (0.1771 peak) \\
\bottomrule
\end{tabular}
\caption{Global scorer behavior during Federated Aesthetic Preference Learning
over 20 rounds.}
\label{tab:app_global_scorer}
\end{table}

\begin{table}
\centering
\scriptsize
\setlength{\tabcolsep}{4.0pt}
\begin{tabular}{@{}clccc@{}}
\toprule
Support & Scorer initialization & SRCC$\uparrow$ & PLCC$\uparrow$ & MSE$\downarrow$ \\
\midrule
10 & Centralized & 0.5291 & 0.5404 & 0.0532 \\
10 & Federated, fixed HP & 0.5412 & 0.5471 & 0.0589 \\
10 & Federated, per-user HPO & 0.5413 & 0.5375 & 0.0700 \\
\midrule
100 & Centralized & 0.5690 & 0.5727 & 0.0552 \\
100 & Federated, fixed HP & 0.5649 & 0.5665 & 0.0595 \\
100 & Federated, per-user HPO & 0.5625 & 0.5646 & 0.0609 \\
\bottomrule
\end{tabular}
\caption{Mean test results after Personalized Scorer Calibration on unseen
users.}
\label{tab:app_personalized_scorer}
\end{table}

With 10 support ratings, the federated initialization with fixed calibration
hyperparameters gives higher SRCC
and PLCC than the centralized counterpart, while the centralized model has
lower MSE. With 100 ratings, the centralized model is marginally strongest on
all three metrics. Both federated variants improve their correlations when the
support set increases from 10 to 100 ratings. Fixed-HP calibration remains
competitive with per-user scorer HPO in both regimes.
For the federated initialization with fixed calibration hyperparameters, the user-level SRCC standard deviations
are 0.130 and 0.116 in the 10- and 100-shot settings, respectively.

\subsection{Generic Enhancement Prior Learning}

Tab.~\ref{tab:app_global_clut} reports a training-target comparison for five
expert-specific CLUT models and one mixed-style model. Each expert-specific
model is evaluated against that expert's retouches, while the mixed-style
configuration uses its associated retouch targets. Because the target
distributions differ across rows, these results characterize target-specific
reconstruction behavior rather than a common-reference ranking.

\begin{table}
\centering
\scriptsize
\setlength{\tabcolsep}{8pt}
\begin{tabular}{@{}lc@{}}
\toprule
CLUT training target & PSNR (dB)$\uparrow$ \\
\midrule
Expert A & 22.28 \\
Expert B & 27.36 \\
Expert~C & 25.21 \\
Expert D & 24.06 \\
Expert E & 25.31 \\
Mixed-style & 23.69 \\
\bottomrule
\end{tabular}
\caption{Training-target comparison for expert-specific and mixed-style CLUT
models under their respective retouch targets.}
\label{tab:app_global_clut}
\end{table}

\subsection{Enhancer Adaptation Strategy Comparison}

Tab.~\ref{tab:app_enhancer_strategy} compares three adaptation and
checkpoint-selection strategies: absolute-score selection, fixed-HP gain
selection, and shared-HPO regularized selection. Input-reference metrics
measure preservation rather than absolute enhancement quality. For the first
two strategies, the displayed score and fidelity values come from test data,
whereas the bracketed preference gains are validation quantities used during
selection. The shared-HPO strategy reports both proxy quantities on held-out
test images. The superscripts make each quantity's evaluation role explicit,
and the analysis compares checkpoint-selection behavior across the three
strategies.

\begin{table}
\centering
\scriptsize
\setlength{\tabcolsep}{2.0pt}
\begin{tabular}{@{}clcc@{}}
\toprule
Support & Strategy & Score [$\Delta$]$\uparrow$ & $P/S/L$-in \\
\midrule
10 & Absolute-score selection & $0.5483^{t}$ [$0.0083^{v}$] & 27.30/0.9515/0.0308 \\
10 & Fixed-HP gain selection & $0.5039^{t}$ [$0.0247^{v}$] & 30.43/0.9627/0.0162 \\
10 & Shared HPO + reg. selection & $0.5624^{t}$ [$0.0826^{t}$] & 20.51/0.7665/0.1206 \\
\midrule
100 & Absolute-score selection & $0.5519^{t}$ [$-0.0004^{v}$] & 29.18/0.9643/0.0234 \\
100 & Fixed-HP gain selection & $0.5245^{t}$ [$0.0244^{v}$] & 31.12/0.9716/0.0132 \\
100 & Shared HPO + reg. selection & $0.5864^{t}$ [$0.0858^{t}$] & 19.76/0.7419/0.1348 \\
\bottomrule
\end{tabular}
\caption{Comparison of enhancer-adaptation and checkpoint-selection
strategies. Superscripts $v$ and $t$ denote validation- and test-split proxy
values. Fidelity columns use the input as reference, and $P/S/L$ denotes
PSNR/SSIM/LPIPS. Bracketed values are preference gains, not standard deviations.
The shared-HPO configuration is selected on validation partitions and evaluated
on held-out test images.}
\label{tab:app_enhancer_strategy}
\end{table}

Absolute-score selection yields smaller validation preference gains and lower
input fidelity than fixed-HP gain selection in both support regimes, indicating
that absolute score alone is a weaker selection signal for user-specific
improvement. Fixed-HP gain selection is the strict open-world reference: it
preserves the input closely and produces a positive validation preference gain
with both support sizes. Shared-HPO regularized selection demonstrates stronger
optimization of the scorer proxy and permits a stronger transformation,
occupying a more preference-oriented operating point. With 100 ratings,
fixed-HP gain selection improves
input-reference PSNR from 30.43 to 31.12~dB, increases SSIM from 0.9627 to
0.9716, and reduces LPIPS from 0.0162 to 0.0132. Shared-HPO regularized
selection raises the mean test score from 0.5624 to 0.5864 and the mean test
preference gain from 0.0826 to 0.0858. These aggregate changes support the value
of additional ratings while preserving the distinction between fixed-HP
fidelity and shared-HPO proxy-preference strength.

\subsection{Detailed Enhancement Baselines and Protocol Context}

The main paper presents the enhancement comparison in a compact consolidated
table. Here we separate results obtained under the common evaluation protocol
from literature-reported operating points so that the source of every number
remains explicit. Tab.~\ref{tab:app_controlled_baseline} is the controlled
comparison: all rows use the same Expert~C reference and evaluation pipeline.

\begin{table}
\centering
\scriptsize
\setlength{\tabcolsep}{5pt}
\begin{tabular}{@{}clccc@{}}
\toprule
Support & Enhancer & PSNR-C$\uparrow$ & SSIM-C$\uparrow$ & LPIPS-C$\downarrow$ \\
\midrule
10 & Fixed HP & 17.97 & 0.801 & 0.130 \\
10 & HPO & 18.33 & 0.777 & 0.147 \\
100 & Fixed HP & 18.12 & 0.804 & 0.129 \\
100 & HPO & 18.58 & 0.775 & 0.147 \\
\bottomrule
\end{tabular}
\caption{FedPAIE evaluated against Expert~C retouches, which provide a
standardized generic professional reference.}
\label{tab:app_expert_c}
\end{table}

\begin{table}
\centering
\scriptsize
\setlength{\tabcolsep}{2.5pt}
\begin{tabular}{@{}lccc@{}}
\toprule
Method & PSNR-C$\uparrow$ & SSIM-C$\uparrow$ & LPIPS-C$\downarrow$ \\
\midrule
Original input & 17.84 & 0.791 & 0.138 \\
AdaInt + personalized scorer~\cite{yang2022adaint} & 19.45 & 0.756 & 0.231 \\
FedPAIE, 10-shot fixed HP & 17.97 & 0.801 & 0.130 \\
FedPAIE, 100-shot fixed HP & 18.12 & 0.804 & 0.129 \\
\bottomrule
\end{tabular}
\caption{Controlled Expert~C comparison under the common evaluation protocol.
AdaInt has the highest PSNR, while 100-shot FedPAIE has the highest SSIM and
lowest LPIPS.}
\label{tab:app_controlled_baseline}
\end{table}

Relative to personalized AdaInt, 10- and 100-shot FedPAIE increase SSIM by
0.045 and 0.048 (approximately 6.0\% and 6.3\%) and reduce LPIPS by 0.101 and
0.102 (approximately 43.7\% and 44.2\%). AdaInt retains PSNR advantages of
1.48 and 1.33~dB. Because all three metrics use Expert~C as a generic
professional reference, these differences characterize a fidelity trade-off
rather than direct evidence of user-preference superiority.

Tab.~\ref{tab:app_literature_context} complements the controlled comparison
with representative personalized enhancement results reported in prior work.
The rows retain their original datasets, targets, resolutions, and evaluation
implementations. They therefore characterize the broader quality--efficiency
landscape, while Tab.~\ref{tab:app_controlled_baseline} provides the direct
method comparison. We omit unreported or speculative latency estimates.
For SpliNet, PSNR and SSIM follow the 20-preference FiveK evaluation reported
by Kosugi and Yamasaki~\cite{kosugi2024personalized}. Its 0.03M parameter count
is computed from the official 10-node, eight-base-filter personalized
architecture~\cite{bianco2020personalized}, which contains approximately
31.4K trainable parameters, rather than quoted from the original paper.
SpliNet is therefore smaller than FedPAIE in raw trainable-parameter count.
The two models have different transformation and system scopes: SpliNet
predicts global per-channel neural-spline color transforms, whereas FedPAIE
uses an image-adaptive 3D-LUT enhancer in a pipeline that connects Federated
Aesthetic Preference Learning with local user adaptation. The 0.293M FedPAIE
figure thus describes a lightweight image-adaptive enhancer in a broader
federated personalization pipeline, while SpliNet remains the smallest architecture in
Tab.~\ref{tab:app_literature_context}.

\begin{table}
\centering
\scriptsize
\setlength{\tabcolsep}{4.2pt}
\begin{tabular}{@{}lccc@{}}
\toprule
Literature-reported method & PSNR & SSIM & Parameters \\
\midrule
SpliNet~\cite{bianco2020personalized} & 18.74 & 0.819 & 0.03M \\
PieNet~\cite{kim2020pienet} & 20.52 & 0.850 & 28M \\
Masked Style Modeling~\cite{kosugi2024personalized} & 22.98 & 0.897 & 90M \\
\bottomrule
\end{tabular}
\caption{Literature-reported personalized enhancement results under their
original evaluation protocols. The table provides broader quality and
model-scale context for the controlled comparison in
Tab.~\ref{tab:app_controlled_baseline}.}
\label{tab:app_literature_context}
\end{table}

\section{Ablation, Qualitative, and Image-Suitability Analyses}
\label{app:analyses}

\subsection{Objective Ablation for Frozen-Scorer-Guided Enhancer Adaptation}

This ablation acts only on Frozen-Scorer-Guided Enhancer Adaptation. It does not
change Federated Aesthetic Preference Learning or Personalized Scorer
Calibration. The 10- and 100-shot studies contain 36 and 37 eligible unseen
users, respectively. Within each support regime, all variants use the same FiveK
evaluation pairs, frozen personalized scorers, shared enhancer HPO
configuration selected for the Full objective, validation-SRCC eligibility
rule, hash-verified original-score cache, training schedule, preprocessing, and
checkpoint-resume policy. Eligibility is
determined only from validation SRCC. Held-out test SRCC does not affect client
inclusion. No removal variant re-runs shared enhancer HPO, and every retained coefficient is copied
from Tab.~\ref{tab:app_enhancer_hparams}. Fixing the shared configuration across
all variants yields a matched within-cohort intervention in which the sole
change is which terms of the Personalized Enhancement Objective remain active:

\begin{table}
\centering
\scriptsize
\setlength{\tabcolsep}{3.5pt}
\begin{tabular}{@{}lccccc@{}}
\toprule
Variant & $\mathcal L_{\mathrm{pref}}$ & $\mathcal L_{\mathrm{aes}}$ &
$\mathcal L_1$ & $\mathcal L_{\mathrm{perc}}$ & $\mathcal L_{\mathrm{gap}}$ \\
\midrule
Full objective & Yes & Yes & Yes & Yes & Yes \\
Without reg. group & Yes & Yes & -- & -- & -- \\
Without scorer guidance & -- & -- & Yes & Yes & -- \\
Without excess-gap penalty & Yes & Yes & Yes & Yes & -- \\
Without $\mathcal L_{\mathrm{pref}}$ & -- & Yes & Yes & Yes & Yes \\
\bottomrule
\end{tabular}
\caption{Active objective terms for Frozen-Scorer-Guided Enhancer Adaptation.
``Without reg. group'' removes the fidelity-plus-gap regularization group
$(\mathcal L_1,\mathcal L_{\mathrm{perc}},\mathcal L_{\mathrm{gap}})$.}
\label{tab:app_ablation_design}
\end{table}

The evaluation includes two reusable controls that require no additional
training. Original is the unenhanced input, and Generic Enhancement Prior
denotes the shared pretrained CLUT-Net checkpoint obtained through Generic
Enhancement Prior Learning without personalization. The Full
objective and four removal variants use 500 common MIT-Adobe FiveK images with Expert~C
input--target pairs for every eligible user. Each user's frozen personalized
scorer supplies the proxy axis. PSNR, SSIM, and LPIPS relative to Expert~C
supply the fidelity axis. Reporting both axes jointly characterizes preference
improvement and image preservation at each operating point.

\begin{table*}[t]
\centering
\scriptsize
\setlength{\tabcolsep}{3.0pt}
\begin{tabular}{@{}lccccc@{\hspace{8pt}}ccccc@{}}
\toprule
& \multicolumn{5}{c}{\textit{10-shot ($n=36$)}} &
\multicolumn{5}{c}{\textit{100-shot ($n=37$)}} \\
\cmidrule(lr){2-6}\cmidrule(lr){7-11}
Variant & Score [$\Delta$] & Win & $P_C$ & $S_C$ & $L_C$ &
Score [$\Delta$] & Win & $P_C$ & $S_C$ & $L_C$ \\
\midrule
Original
& 0.4324 [0.0000] & -- & -- & -- & --
& 0.4534 [0.0000] & -- & -- & -- & -- \\
Generic prior
& 0.4939 [+0.0615] & 100 & 22.60 & 0.904 & 0.087
& 0.5153 [+0.0619] & 100 & 22.60 & 0.904 & 0.087 \\
Full objective
& 0.5213 [+0.0890] & 100 & 18.33 & 0.777 & 0.147
& 0.5438 [+0.0904] & 100 & 18.58 & 0.775 & 0.147 \\
w/o $\mathcal L_{\mathrm{pref}}$
& 0.5160 [+0.0836] & 100 & 18.46 & 0.789 & 0.139
& 0.5436 [+0.0903] & 100 & 18.45 & 0.775 & 0.145 \\
w/o $\mathcal L_{\mathrm{gap}}$
& 0.5295 [+0.0971] & 100 & 13.43 & 0.495 & 0.341
& 0.5513 [+0.0979] & 100 & 14.23 & 0.534 & 0.310 \\
w/o reg. group
& 0.5173 [+0.0850] & 94 & 7.30 & 0.235 & 0.548
& 0.5269 [+0.0736] & 86 & 7.73 & 0.289 & 0.529 \\
w/o scorer guidance
& 0.4326 [+0.0002] & 81 & 18.05 & 0.814 & 0.122
& 0.4535 [+0.0001] & 68 & 18.04 & 0.814 & 0.122 \\
\bottomrule
\end{tabular}
\caption{Complete dual-axis objective ablation on 500 FiveK images. Score is
the mean frozen personalized scorer output, and $\Delta$ is the scorer-predicted
preference gain from
Original, and the win rate is the fraction of users with a positive mean
change. Personalized rows aggregate $36\times500$ or $37\times500$ outputs.
The Generic Enhancement Prior fidelity metrics are computed once over the 500
images and shared across users. Expert~C provides a standardized professional
reference.
Displayed statistics are rounded independently from unrounded aggregates.
Bracketed values are gains, not standard deviations.}
\label{tab:app_ablation_results}
\end{table*}

\begin{figure*}[t]
\centering
\includegraphics[width=0.49\textwidth]{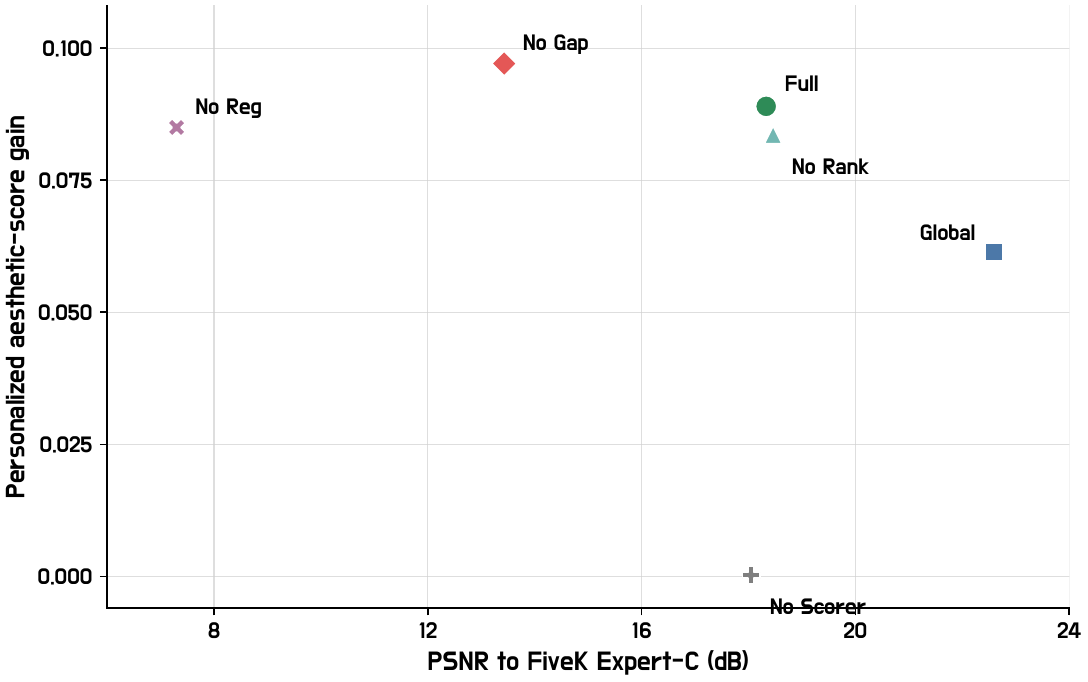}\hfill
\includegraphics[width=0.49\textwidth]{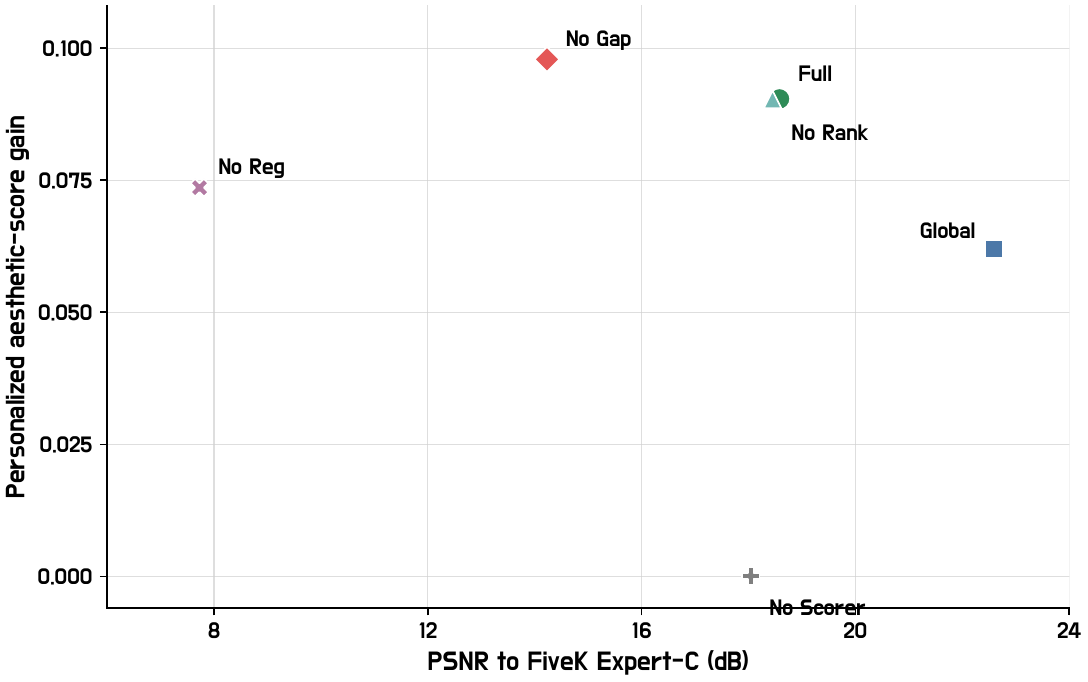}
\caption{Scorer-predicted preference-gain--fidelity trade-off in the objective
ablation for Frozen-Scorer-Guided Enhancer Adaptation on 500 FiveK images.
The left and right panels show the 10-shot and 100-shot settings,
respectively. Each labeled point relates mean gain from Original under the
same frozen personalized scorer to PSNR relative to Expert~C. Global denotes
the Generic Enhancement Prior, No Rank denotes w/o
$\mathcal L_{\mathrm{pref}}$, No Gap denotes w/o
$\mathcal L_{\mathrm{gap}}$, No Reg denotes w/o the fidelity-plus-gap
regularization group, and
No Scorer denotes w/o scorer guidance. Full improves the scorer proxy beyond
the Generic Enhancement Prior while retaining substantially higher fidelity
than No Gap and No Reg. The score is an optimization-aligned user-specific
proxy, and Expert~C provides a standardized professional reference.}
\label{fig:app_ablation_tradeoff}
\end{figure*}

\par\noindent\textbf{Personalization Relative to the Controls.}
Tab.~\ref{tab:app_ablation_results} and
Fig.~\ref{fig:app_ablation_tradeoff} show that the Full objective improves
the frozen personalized scorer output over Original by 0.0890 and 0.0904 for 10 and
100 support ratings. It also improves over the Generic Enhancement Prior by
0.0274 and 0.0285. Tab.~\ref{tab:app_ablation_significance} confirms these
differences with paired user-level tests. The Full objective exceeds the
Generic Enhancement Prior for every evaluated user, namely 36 of 36 in the
10-shot setting and 37 of 37 in the 100-shot setting. This universal direction
under the frozen personalized scorer proxy provides stronger user-level
evidence than an aggregate mean alone. Together with the fidelity axes and
qualitative views, it supports the same preference--fidelity operating point.
Fig.~\ref{fig:app_qualitative_ablation_additional} provides two additional
100-shot examples.

\begin{table}
\centering
\scriptsize
\setlength{\tabcolsep}{3.0pt}
\begin{tabular}{@{}clrrr@{}}
\toprule
Support & Comparison & Mean diff. & $t$ & $p$ \\
\midrule
10 & Prior--Original & 0.0615 & 19.01 & $5.0\times10^{-20}$ \\
10 & Full--Original & 0.0890 & 12.43 & $2.1\times10^{-14}$ \\
10 & Full--Prior & 0.0274 & 5.49 & $3.6\times10^{-6}$ \\
100 & Prior--Original & 0.0619 & 26.65 & $2.7\times10^{-25}$ \\
100 & Full--Original & 0.0904 & 19.37 & $1.3\times10^{-20}$ \\
100 & Full--Prior & 0.0285 & 7.97 & $1.8\times10^{-9}$ \\
\bottomrule
\end{tabular}
\caption{Paired user-level tests on frozen personalized scorer outputs. The
tests quantify consistency across users under the learned personalization
proxy. Prior denotes the Generic Enhancement Prior. The support cohorts contain
36 and 37 paired users.}
\label{tab:app_ablation_significance}
\end{table}
\par\noindent\textbf{Scorer Guidance Supplies the Personalization Signal.}
Without scorer guidance, only $\mathcal L_1+\mathcal L_{\mathrm{perc}}$ remain.
The mean gain becomes 0.0002 in the 10-shot setting and 0.0001 in the 100-shot
setting. On Flickr-AES, the same variant reaches 74.1 and 73.6~dB PSNR relative to
the input with SSIM of approximately 0.9999. It therefore converges to an
identity-like transformation. The 81\% and 68\% positive-direction rates in
Tab.~\ref{tab:app_ablation_results} correspond to minute changes around zero,
not meaningful personalization. This variant provides a direct lower bound and
isolates the frozen personalized scorer as the source of measurable
proxy-guided adaptation.
\par\noindent\textbf{The Excess-Gap Penalty Protects Fidelity.}
Removing only $\mathcal L_{\mathrm{gap}}$ raises the proxy score from 0.5213 to
0.5295 for 10-shot and from 0.5438 to 0.5513 for 100-shot. Read alone, these
numbers would incorrectly favor the removal variant. The fidelity axis reveals the
failure mode. Relative to the Full objective, PSNR falls by 4.90 and 4.35~dB,
SSIM falls by 0.282 and 0.241, and LPIPS increases by 0.194 and 0.163. This
controlled result gives $\mathcal L_{\mathrm{gap}}$ a clear component-level
interpretation. It limits gains that exploit the scorer proxy and preserves a
usable fidelity floor.
\par
\begin{figure*}[t]
\centering
\includegraphics[width=0.92\textwidth]{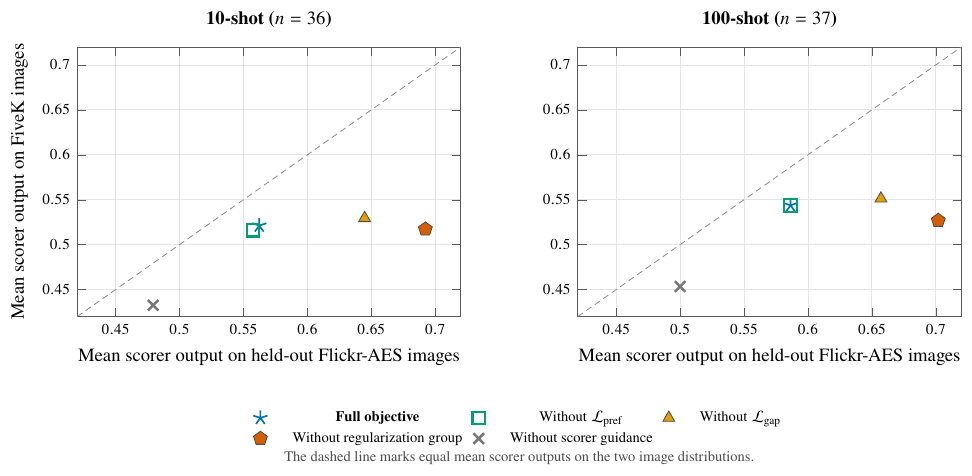}
\caption{Cross-distribution analysis for the objective ablation of
Frozen-Scorer-Guided Enhancer Adaptation. Each point compares the mean frozen personalized scorer output on
held-out Flickr-AES images with that on the common 500-image FiveK set. The
dashed line marks equal mean outputs on the two image distributions. The Full
objective remains nearer this line than the variants without
$\mathcal L_{\mathrm{gap}}$ or the complete fidelity-plus-gap regularization group. Their larger
decreases, together with the low input-reference fidelity in
Tab.~\ref{tab:app_cross_distribution}, are consistent with non-transferable
proxy over-optimization. Together with the paired tests and fidelity metrics,
this analysis provides complementary evidence across image distributions.}
\label{fig:app_ablation_cross_distribution}
\end{figure*}

\noindent\textbf{The Fidelity-Plus-Gap Regularization Group Improves Cross-Distribution Behavior.}
Jointly removing $\mathcal L_1$, $\mathcal L_{\mathrm{perc}}$, and
$\mathcal L_{\mathrm{gap}}$ produces the highest in-domain Flickr-AES scores,
0.6923 and 0.7019, but simultaneously reduces input-reference PSNR to 7.42 and
7.95~dB. On FiveK, its scores fall below the Full objective while its
PSNR relative to Expert~C reaches only 7.30 and 7.73~dB. The apparent in-domain
advantage therefore does not transfer. Full improves every reported proxy and
fidelity measure over this removal variant. It restores 11.03 and 10.85~dB PSNR
and reduces LPIPS by 0.401 and 0.382. Tab.~\ref{tab:app_cross_distribution}
shows the full comparison. These results support the fidelity-plus-gap regularization group as a
joint defense against proxy exploitation. The comparison between the variant
without $\mathcal L_{\mathrm{gap}}$ and the variant without the complete
fidelity-plus-gap regularization group further isolates the two fidelity losses as a pair.
Retaining $\mathcal L_1+\mathcal L_{\mathrm{perc}}$ improves FiveK PSNR by
6.13 and 6.50~dB, increases SSIM by 0.260 and 0.245, and reduces LPIPS by
0.207 and 0.219. It also raises the FiveK proxy score by 0.0122 and 0.0244.
Thus, the two fidelity losses jointly improve both the usable operating point
and cross-distribution stability rather than merely suppressing enhancement.
The paired intervention evaluates $\mathcal L_1$ and
$\mathcal L_{\mathrm{perc}}$ as the intended fidelity component, matching
their joint role in preserving pixel and perceptual content. The
Flickr-AES-to-FiveK score decrease is 0.0411 and
0.0426 for Full, compared with 0.1153 and 0.1058 without
$\mathcal L_{\mathrm{gap}}$ and 0.1750 in both regimes without the complete
fidelity-plus-gap regularization group. This gap provides an additional transfer-oriented view
of the protective terms. Fig.~\ref{fig:app_ablation_cross_distribution}
visualizes the same controlled cross-distribution comparison.

\begin{table}
\centering
\scriptsize
\setlength{\tabcolsep}{2.5pt}
\begin{tabular}{@{}
>{\raggedright\arraybackslash}p{0.31\columnwidth}
>{\centering\arraybackslash}p{0.32\columnwidth}
>{\centering\arraybackslash}p{0.28\columnwidth}@{}}
\toprule
Variant &
\shortstack{Flickr score / $P_{\mathrm{in}}$\\10-shot / 100-shot} &
\shortstack{FiveK score\\10-shot / 100-shot} \\
\midrule
Full objective &
\shortstack{0.5624/20.51\\0.5864/19.76} &
\shortstack{0.5213\\0.5438} \\
w/o $\mathcal L_{\mathrm{pref}}$ &
\shortstack{0.5575/21.10\\0.5864/19.90} &
\shortstack{0.5160\\0.5436} \\
w/o $\mathcal L_{\mathrm{gap}}$ &
\shortstack{0.6448/13.30\\0.6571/14.44} &
\shortstack{0.5295\\0.5513} \\
w/o reg. group &
\shortstack{0.6923/7.42\\0.7019/7.95} &
\shortstack{0.5173\\0.5269} \\
w/o scorer guidance &
\shortstack{0.4794/74.1\\0.5001/73.6} &
\shortstack{0.4326\\0.4535} \\
\bottomrule
\end{tabular}
\caption{Cross-distribution analysis. Flickr-AES values use held-out
in-domain test images, and PSNR-in measures change from the input. FiveK
scores use the 500-image evaluation set. Each stacked cell lists the 10-shot
value above the 100-shot value. A high scorer output
accompanied by very low input fidelity and weak transfer is consistent with
proxy exploitation.}
\label{tab:app_cross_distribution}
\end{table}

\par\noindent\textbf{The Preference Terms Are Compatible.}
Removing $\mathcal L_{\mathrm{pref}}$ reduces the 10-shot score from 0.5213 to
0.5160, while the 100-shot score changes from 0.5438 to 0.5436. Fidelity stays
close in both regimes. At the selected weights,
$\mathcal L_{\mathrm{aes}}$ already provides a strong absolute preference
signal, while $\mathcal L_{\mathrm{pref}}$ explicitly encodes the desired
ordering between enhanced and original images. Its additional effect is most
visible under sparse support, while preserving the 100-shot solution. The two preference-bearing terms therefore provide
overlapping and compatible supervision.
\par\noindent\textbf{Why the Generic Enhancement Prior Has Higher Fidelity to Expert~C.}
The Generic Enhancement Prior is initialized from paired FiveK professional
retouches, whereas personalized adaptation deliberately moves its output
toward each user's frozen personalized scorer. The Expert~C axis is therefore naturally
favorable to the generic professional prior. Its PSNR of 22.60~dB should be
read together with the Full objective's higher personalized score and
consistent per-user advantage. The result characterizes the intended
personalization--reference trade-off.

\subsection{Orthogonal Optimization-Choice Ablation}

The loss-component ablation above fixes the optimization configuration and
changes only the active objective terms. We separately vary enhancer
hyperparameters, Personalized Scorer Calibration, and support size. These
factors form a $2\times2\times2$ design and answer a different question from
component attribution.

\begin{table}
\centering
\scriptsize
\setlength{\tabcolsep}{0.5pt}
\begin{tabular}{@{}llcc@{}}
\toprule
Enhancer HP & Scorer calibration &
10-shot score / PSNR-in & 100-shot score / PSNR-in \\
\midrule
Fixed HP & Fixed HP & 0.499 / 28.4 & 0.510 / 30.8 \\
Fixed HP & \shortstack{Per-user scorer\\HPO} & 0.504 / 30.4 & 0.525 / 31.1 \\
\shortstack{Shared enhancer\\HPO} & Fixed HP & 0.547 / 18.5 & 0.572 / 19.3 \\
\shortstack{Shared enhancer\\HPO} & \shortstack{Per-user scorer\\HPO} & 0.562 / 20.5 & 0.586 / 19.8 \\
\bottomrule
\end{tabular}
\caption{Orthogonal $2\times2\times2$ optimization-choice ablation. Scores and
PSNR-in are measured on the Flickr-AES test split, with the input as the PSNR
reference. HP and HPO denote hyperparameters and hyperparameter optimization.}
\label{tab:app_design_choice_ablation}
\end{table}

Tab.~\ref{tab:app_design_choice_ablation} shows that per-user scorer HPO in
Personalized Scorer Calibration improves the predicted score under both
enhancer configurations and both support sizes. It also improves
input-reference PSNR in every matched comparison. Shared enhancer HPO produces the
larger score increase and intentionally permits a stronger transformation,
which lowers PSNR-in relative to the fixed-HP configuration. Combining the two choices gives the
highest frozen personalized scorer output at both support sizes while recovering some
fidelity relative to shared enhancer HPO with fixed-HP scorer calibration.

\begin{figure*}[t]
\centering
\includegraphics[width=\textwidth]{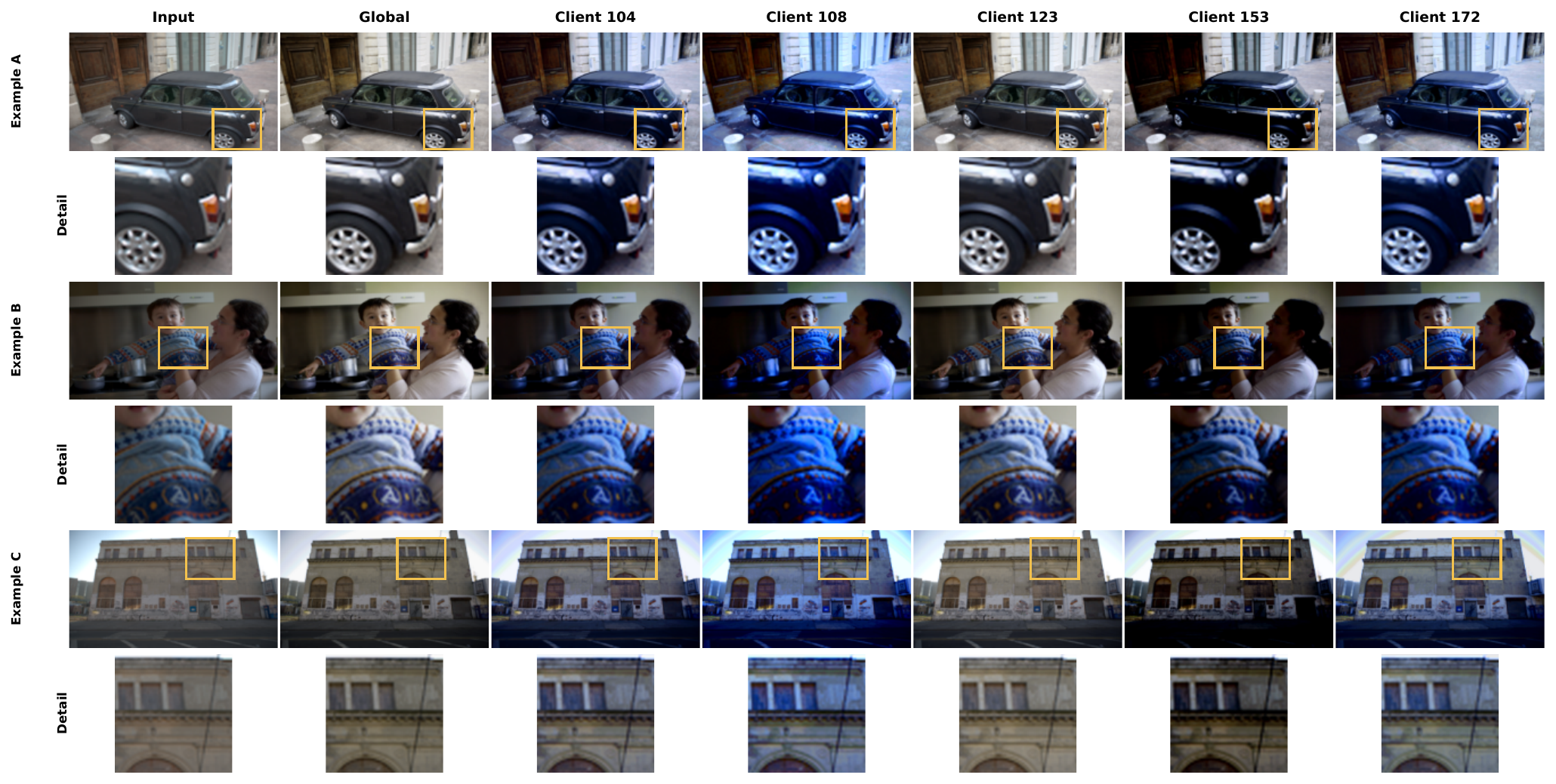}
\caption{Full-image and detail-crop comparison for Examples~A--C. Yellow boxes
mark the enlarged regions. The crops make client-dependent tonal and chromatic
shifts visible around the wheel, clothing texture, and building facade while
showing that local structures remain aligned.}
\label{fig:app_detail}
\end{figure*}

This factorial result supports the selected configuration without conflating
optimization choices with loss-component evidence. Results for Federated
Aesthetic Preference Learning and Personalized Scorer Calibration remain in
Tab.~\ref{tab:app_personalized_scorer}, where the federated initialization with
fixed-HP calibration
reaches SRCC $0.541\pm0.130$ and $0.565\pm0.116$ for 10 and
100 support ratings. Together, these experiments evaluate the complete
Lightweight Dual-Cue Aesthetic Scorer under both sparse and richer support.

\subsection{Qualitative and Client-Level Analyses}

\begin{figure}[t]
\centering
\includegraphics[width=\columnwidth]{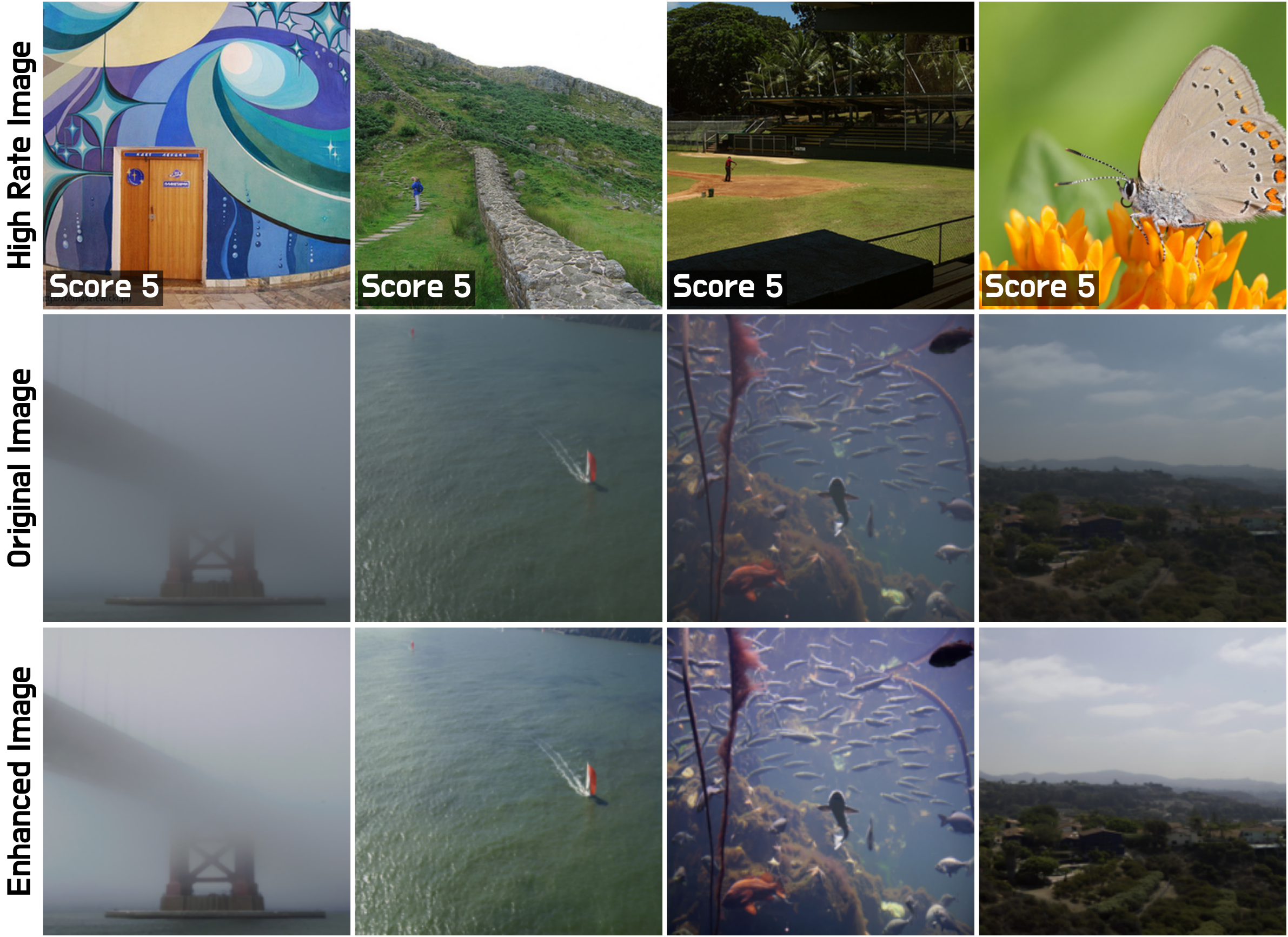}
\caption{Additional preference context and Full-objective outputs for
Flickr-AES user~26 after 100-shot Personalized Scorer Calibration. The top row
shows four rating-5 Flickr-AES examples, the middle row shows four unpaired
FiveK inputs, and the bottom row shows their outputs after
Frozen-Scorer-Guided Enhancer Adaptation. The rating examples provide user-level
preference context rather than paired target-style supervision.}
\label{fig:app_user26_preference}
\end{figure}

\begin{figure*}[t]
\centering
\includegraphics[width=0.76\textwidth]{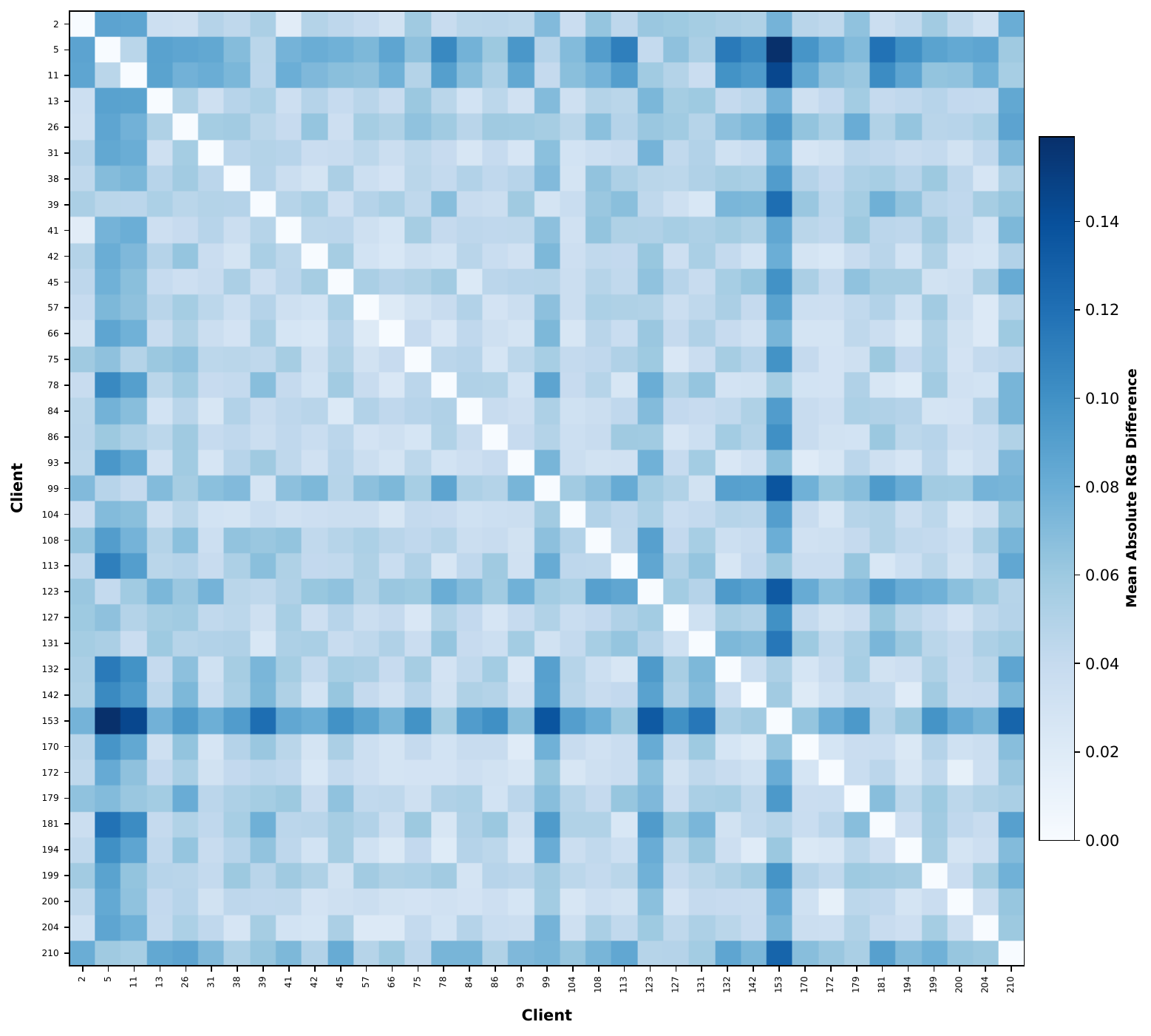}
\caption{Pairwise output diversity over the full 100-shot evaluation cohort.
Each cell is the mean absolute RGB difference between two personalized
enhancers, averaged over the common FiveK inputs. The diagonal is zero by
definition. Larger values indicate stronger client-dependent variation, while
the scorer and fidelity analyses evaluate preference alignment and quality.}
\label{fig:app_all_client_matrix}
\end{figure*}

The main paper presents the complete shared-input comparison. Across
Examples~A--E, Clients~104, 108, and 172 favor cooler outputs, Client~123 stays
closer to the Generic Enhancement Prior, and Client~153 applies a darker,
higher-contrast transform. Fig.~\ref{fig:app_user26_preference} adds a separate
user-level view by placing rating-5 Flickr-AES examples beside unpaired FiveK
inputs and Full-objective outputs. Fig.~\ref{fig:app_detail} shows that tonal and
chromatic differences remain visible in local crops while scene structure is
preserved. Fig.~\ref{fig:app_all_client_matrix} extends the analysis beyond the
displayed clients. The nonuniform pairwise distances show that output diversity
is distributed across the full 100-shot evaluation cohort rather than confined to the
selected qualitative examples. These figures jointly document distinct
transformations and user-specific preference context.

\begin{figure*}[t]
\centering
\includegraphics[width=0.49\textwidth]{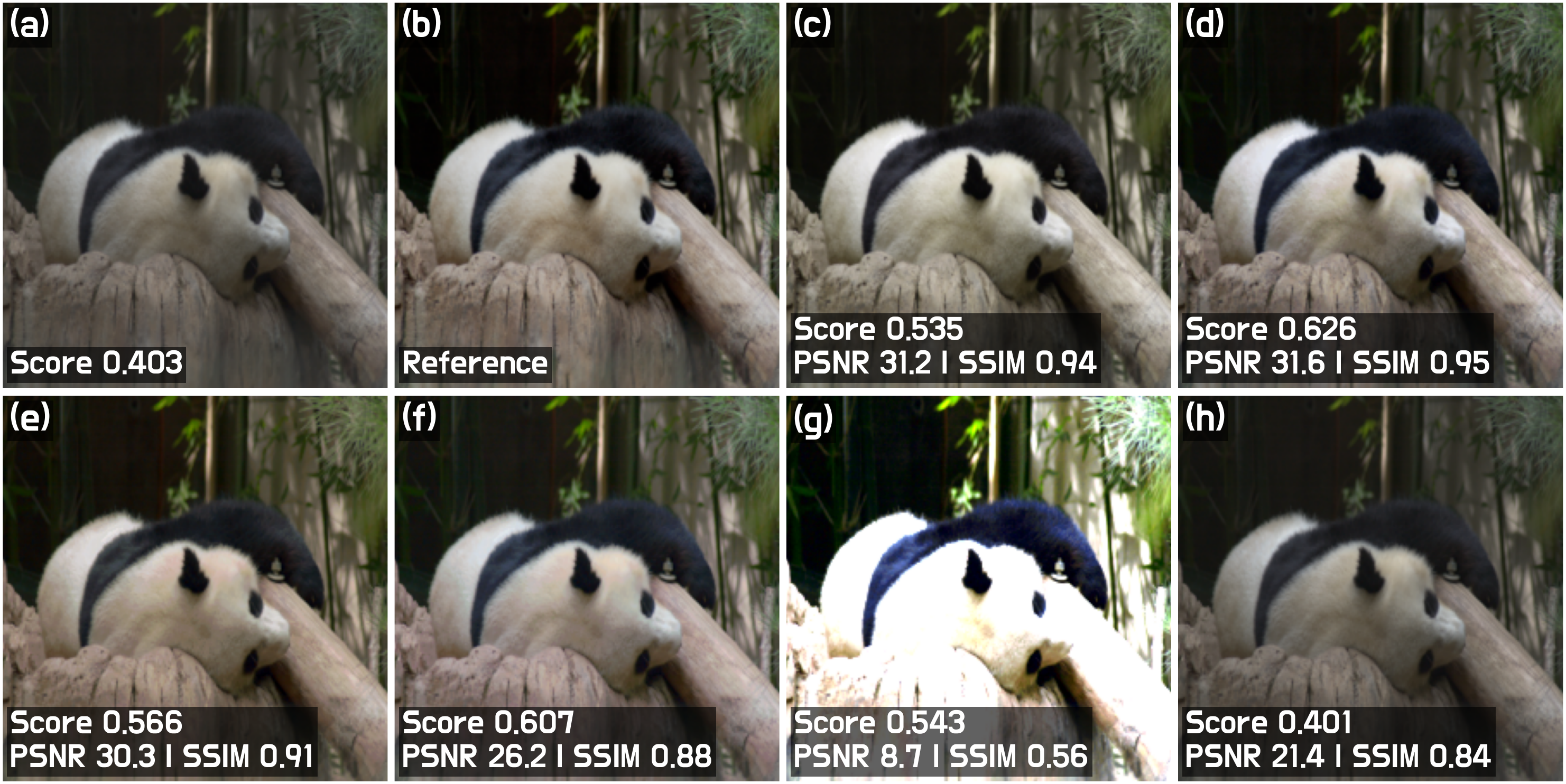}\hfill
\includegraphics[width=0.49\textwidth]{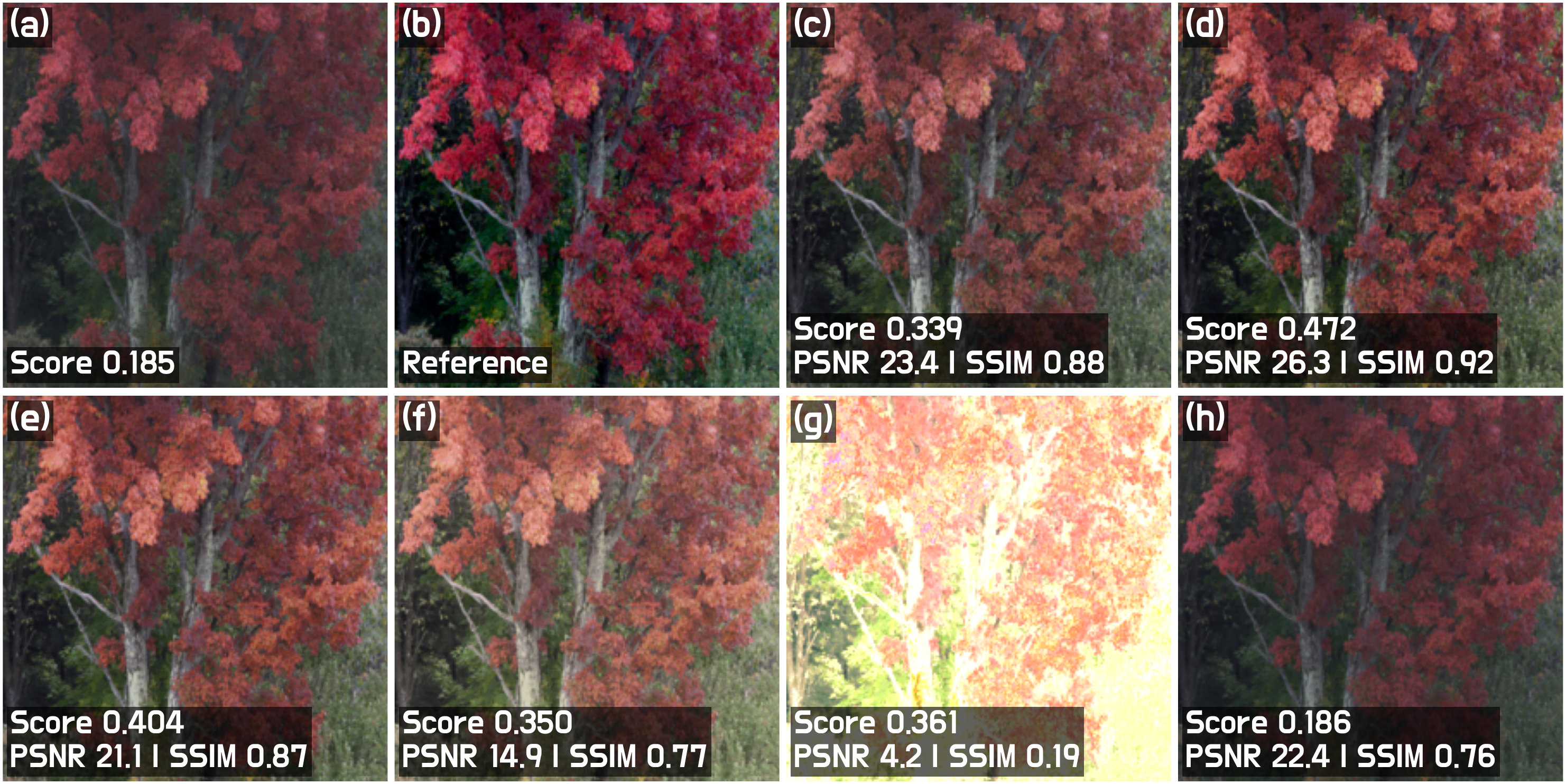}
\caption{Additional 100-shot objective-ablation cases for user~199 (left) and
user~26 (right). Within each case, panels show (a) Original, (b) the Expert~C
reference, (c) the Generic Enhancement Prior, (d) the Full objective, (e)
without $\mathcal L_{\mathrm{pref}}$, (f) without
$\mathcal L_{\mathrm{gap}}$, (g) without the fidelity-plus-gap regularization
group, and (h) without scorer guidance. Scores are frozen personalized scorer
outputs, while PSNR and SSIM use Expert~C as reference. On both displayed
images, Full improves the frozen-scorer output and Expert~C fidelity over the
Generic Enhancement Prior, whereas removing the full regularization group
causes severe overexposure.}
\label{fig:app_qualitative_ablation_additional}
\end{figure*}

\begin{figure*}[p]
\centering
\includegraphics[width=0.76\textwidth]{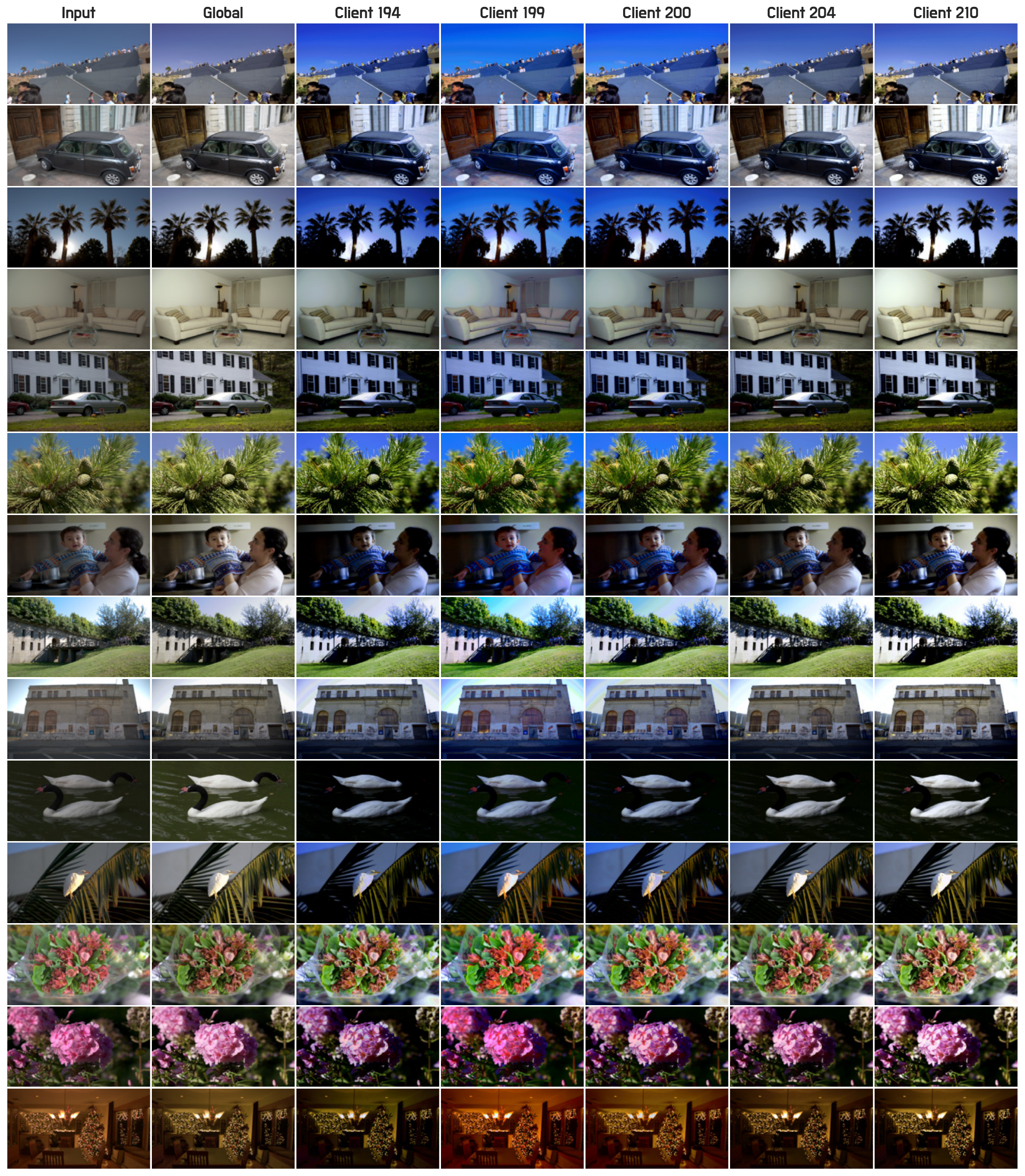}
\caption{Randomly selected shared-input comparison for 100-shot Clients~194,
199, 200, 204, and 210. Each row contains a FiveK input, the Generic Enhancement
Prior output, and five personalized outputs. Python's
\texttt{random.Random(42)} selects one of three pre-generated, non-overlapping
groups. Stable client-dependent color transformations remain visible across
the shared inputs.}
\label{fig:app_shared_inputs_random}
\end{figure*}

To complement the main-paper user~204 example, the additional visualizations
use users~199 and~26. On the displayed images, Full raises the frozen-scorer
output from $0.535$ to $0.626$ and from $0.339$ to $0.472$ relative to the
Generic Enhancement Prior, while PSNR/SSIM increase from $31.2/0.94$ to
$31.6/0.95$ and from $23.4/0.88$ to $26.3/0.92$, respectively.
Fig.~\ref{fig:app_qualitative_ablation_additional} shows both cases, and
Fig.~\ref{fig:app_user26_preference} provides complementary preference
context.
Fig.~\ref{fig:app_shared_inputs_random} extends the shared-input comparison
with a randomly selected sheet containing five clients. Python's
\texttt{random.Random(42)} selects one of three pre-generated, non-overlapping
candidate cohorts, yielding Clients~194, 199, 200, 204, and 210. Aggregate
claims rely on the complete-cohort tables and matrix.

\subsection{Image-Conditioned Output Variation}

\begin{figure*}[p]
\centering
\includegraphics[width=0.90\textwidth]{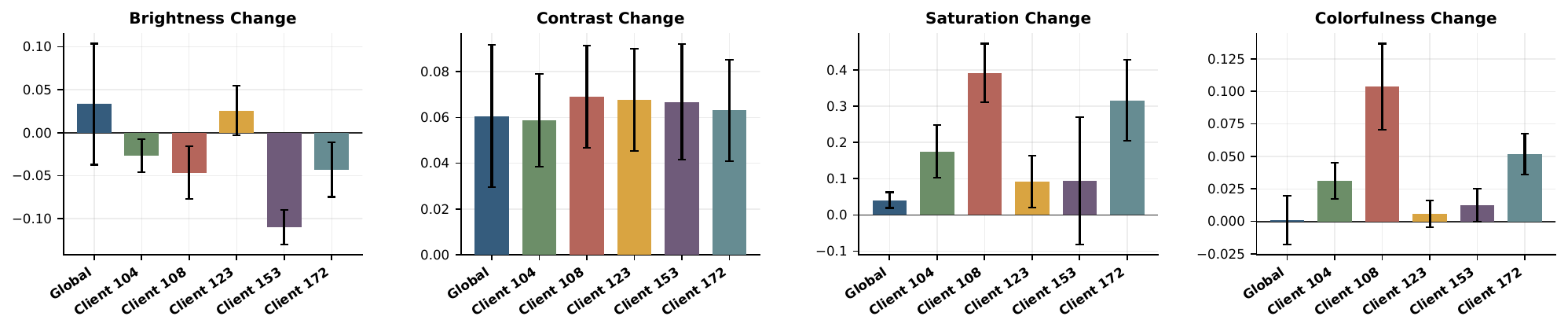}
\caption{Attribute changes from the input for the Generic Enhancement Prior and
five main-paper clients. Bars and error bars show the mean and variation across
qualitative examples. The statistics characterize each client's enhancement
style.}
\label{fig:app_attribute_shift}
\end{figure*}

For a common input $I$ and $U$ personalized enhancers, we define the average
pairwise output difference as:
\begin{equation}
D(I)=\frac{2}{U(U-1)}\sum_{u<v}
\operatorname{MAE}\!\left(E_{\phi_u}(I),E_{\phi_v}(I)\right).
\label{eq:app_interclient_difference}
\end{equation}
Here $u$ and $v$ index personalized users and $\operatorname{MAE}$ is the mean
absolute RGB error between two outputs.
The analysis applies 33 valid 10-shot enhancers to 500 inputs. $D(I)$ ranges
from 0.011 to 0.073. The 15 lowest- and 15 highest-difference inputs form the
LOW and HIGH groups, respectively. Tab.~\ref{tab:app_image_attributes} shows
that the HIGH group is brighter, less saturated, and lower contrast on
average. These statistics characterize the joint attribute profile of the two
groups. Fig.~\ref{fig:app_attribute_shift} complements this input-level
grouping by summarizing how each displayed output changes
brightness, contrast, saturation, and colorfulness relative to the input.

\begin{table}
\centering
\scriptsize
\setlength{\tabcolsep}{7pt}
\begin{tabular}{@{}lccc@{}}
\toprule
Group & Brightness & Saturation & Contrast \\
\midrule
LOW $D(I)$ & 0.167 & 0.402 & 0.145 \\
HIGH $D(I)$ & 0.346 & 0.195 & 0.111 \\
\bottomrule
\end{tabular}
\caption{Mean input attributes for the lowest- and highest-difference groups.}
\label{tab:app_image_attributes}
\end{table}

We additionally compare proxy scores for the LOW and HIGH groups
(Tab.~\ref{tab:app_suitability_proxy}). FedPAIE produces the
largest HIGH--LOW change, 0.0904, exceeding SpliNet by 31.2\% and PieNet by
10.6\%. This result supports the central observation that FedPAIE expresses a
larger image-conditioned proxy response when an input offers more editable color space.
Because identical scorer-calibration and generation protocols are not
established for all three methods, the absolute score levels retain their
method-specific context. The relative response between LOW and HIGH groups is
the informative comparison here.

\begin{table}
\centering
\scriptsize
\setlength{\tabcolsep}{6pt}
\begin{tabular}{@{}lccc@{}}
\toprule
Method & LOW & HIGH & HIGH$-$LOW \\
\midrule
FedPAIE & 0.4404 & 0.5308 & 0.0904 \\
SpliNet~\cite{bianco2020personalized} & 0.4990 & 0.5679 & 0.0689 \\
PieNet~\cite{kim2020pienet} & 0.4983 & 0.5800 & 0.0817 \\
\bottomrule
\end{tabular}
\caption{Reported proxy-score response across image groups. FedPAIE shows the
largest HIGH--LOW change in this analysis.}
\label{tab:app_suitability_proxy}
\end{table}

\section{Evaluation Scope and Evidence Interpretation}
\label{app:limitations}

The experimental suite combines a user-disjoint open-world reference, matched
objective interventions, optimization-aligned personalization measures, and
reference-based fidelity measures. The following scope clarifies how these
complementary forms of evidence support the conclusions.
\par\noindent\textbf{Model Selection.}
Per-user scorer HPO uses support images for fitting and a disjoint validation
split for selection, followed by evaluation on held-out images. Shared enhancer
HPO selects one configuration from validation partitions of eight identities,
with all test images held out. This configuration is fixed for within-cohort
trade-off and objective analyses, while the prespecified fixed-HP results
provide the user-disjoint open-world reference. The two tracks separate
open-world generalization from controlled configuration analysis.
\par\noindent\textbf{Personalization Evidence.}
The frozen personalized scorer supplies both the training signal and the
reported scorer-predicted preference gain, making $\Delta$ a direct,
optimization-aligned user-preference endpoint. Paired user-level tests quantify
the consistency of this gain, while input- and Expert-C-reference metrics,
cross-distribution analysis, and qualitative comparisons provide complementary
evidence for fidelity and transformation stability. Blinded user evaluation
offers a natural extension with direct perceptual feedback. The five-setting
objective ablation for
Frozen-Scorer-Guided Enhancer Adaptation is complete for the 10- and 100-shot
settings and supports term-level conclusions for
$\mathcal L_{\mathrm{pref}}$ and $\mathcal L_{\mathrm{gap}}$ and group-level
conclusions for scorer guidance and fidelity-plus-gap regularization within its matched shared enhancer HPO
analysis. The paired fidelity-loss intervention evaluates $\mathcal L_1$ and
$\mathcal L_{\mathrm{perc}}$ together as their functional preservation group,
while the complete dual-cue scorer is evaluated throughout both support
regimes.
\par\noindent\textbf{References and Matched Cohorts.}
Input-reference PSNR, SSIM, and LPIPS measure preservation. Metrics relative to
Expert~C measure proximity to a standardized professional style, while the
frozen personalized scorer provides the user-specific endpoint. The
experimental design uses fixed, analysis-specific cohorts. The 10-shot
evaluation uses 36 users, the 100-shot evaluation uses 37 users, and the
image-suitability analysis uses 33 valid enhancers. Within each objective
analysis, identical eligibility rules and evaluation data are applied across
all variants. This yields matched within-regime comparisons and descriptive
trends across support scales.
\par\noindent\textbf{Comparison Context and Statistical Evidence.}
The literature rows position FedPAIE among representative recent personalized
enhancement methods under their original protocols, while the study-evaluated
rows provide same-protocol evidence. Paired tests cover the complete matched
user cohorts, and the resource analysis reports parameters, trainable state,
and static GFLOPs throughout the pipeline. Together, these results evaluate
FedPAIE at the model, user, and image levels.
\par\noindent\textbf{Privacy Scope.}
Raw images and ratings remain local throughout the implemented protocol. Only
scorer model updates are exchanged during federated training, establishing a
clear protocol-level raw-data privacy boundary. Secure aggregation and
differential privacy are compatible complementary communication-layer
protections.

\section{Extended Related Work and Novelty Positioning}
\label{app:relatedwork}

The main paper provides a compact account of the most relevant literature. This
section expands that discussion by separating five dimensions that are often
conflated: the aesthetic target, the form of preference supervision, the
location of user data, the supervision used to learn an image transformation,
and the model retained for inference. This separation is important because a
method can be personalized without being federated, federated without producing
an image, or computationally efficient without learning an individual user's
preference.

\subsection{Efficient Image Enhancement and Color Grading}
\label{app:rw_enhancement}

Learning-based photo enhancement commonly estimates color and tone
transformations from input--retouch pairs. MIT-Adobe FiveK established a
standard paired setting with multiple expert renditions of each photograph
\cite{bychkovsky2011learning}. Subsequent systems increased content adaptivity
through bilateral-grid prediction~\cite{gharbi2017deep}, removed strict pairing
through adversarial learning~\cite{chen2018deep}, or incorporated an aesthetic
objective into the enhancement process~\cite{deng2018aesthetic}. These methods
made learned enhancement more flexible, but their target is a generic
enhancement distribution or an expert style. They do not infer the preference
of an unseen user from that user's sparse private ratings.

Image-adaptive 3D lookup tables provide an especially efficient form of global
color grading. Zeng et al.\ learn input-dependent mixtures of basis LUTs
\cite{zeng2022learning}. AdaInt learns nonuniform sampling intervals in color
space~\cite{yang2022adaint}, while CLUT-Net factorizes and compresses the LUT
representation~\cite{zhang2022clutnet}. SVDLUT decomposes spatial-aware lookup
tables to reduce model size and runtime while retaining spatial information
\cite{kim2025svdlut}. These LUT methods focus on efficient generic
enhancement. FedPAIE instead uses a compressed representation as a
preference-driven personalization engine
within a federated pipeline. Generic paired retouches train the shared
initialization. For a new user, the compressed LUT bases remain fixed while the
lightweight coefficient predictor is adapted using a preference signal learned
from local ratings. This design turns sparse private ratings into
user-specific, image-adaptive 3D-LUT transformations without requesting paired
retouches from the user.

\subsection{Personalized Aesthetics Assessment and Preference Learning}
\label{app:rw_assessment}

Generic aesthetic assessors estimate population-level quality or preference.
NIMA predicts an aesthetic rating distribution~\cite{talebi2018nima}, and VILA
uses vision--language pretraining to learn aesthetics from user comments
\cite{ke2023vila}. Personalized image-aesthetics assessment instead models the
deviation of an individual's taste from a population prior
\cite{ren2017personalized}. Representative approaches use collaborative
attention~\cite{wang2018collaborative}, rich user and image
attributes~\cite{yang2022personalized}, few-shot meta-learning
\cite{zhu2022personalized}, graph-based collaboration
\cite{shi2024personalized}, or multi-level transitional contrast learning that
exploits cross-user contrastive information~\cite{yang2024multilevel}.
Task-vector customization combines reusable vectors from generic-aesthetic and
image-quality databases for scalable few-shot personalization and cross-domain
generalization~\cite{yun2024scaling}. PAA+ extends the usual pretraining--fine-tuning
paradigm with continual learning and validates it in a physique-aesthetics
setting~\cite{zhong2025paaplus}. Recent datasets also broaden the assessment
domain. LAPIS provides 11,723 artwork images with aesthetic ratings and rich
image and personal attributes for personalized assessment
\cite{maerten2025lapis}. Human preference comparisons provide a general
mechanism for learning from limited feedback~\cite{christiano2017deep}, while
user-guided reinforcement learning has been applied specifically to
personalized image-aesthetics assessment~\cite{lv2021user}. Together with the
assessment methods above, these works show that personalized preferences can
be learned from limited user feedback. The assessment methods themselves
primarily output scores, rankings, or preference representations rather than
enhanced images.

Task-vector customization is especially close to our few-shot calibration
setting, but it builds personalized scorers by centrally combining database
task vectors. FedPAIE instead learns the shared scorer from decentralized
ratings and uses aesthetic assessment as an intermediate model rather than the
terminal task. Its lightweight dual-cue scorer combines color statistics with
semantic context, which makes the prediction sensitive to both the appearance
change and the image content. The global scorer is learned from decentralized
image--rating pairs, after which the Support-Dependent Scorer Mask controls
few-shot calibration for an unseen user. Pairwise ordering and variance
preservation are introduced only during this local calibration. The calibrated
scorer is then frozen and used as a differentiable interface between sparse
ratings and an image transformation. This role is distinct from reporting a
personalized score. Gradients pass through the frozen personalized scorer to update the
enhancer, so the preference estimate becomes a local transformation signal
without making the scorer itself drift toward the images it rewards.

\subsection{Personalized Image Enhancement}
\label{app:rw_personalized_enhancement}

Personalized enhancement directly aims to produce different outputs for
different users. Early work learns an individual preference model from user
choices over candidate adjustments~\cite{kang2010personalization}. PieNet
represents a user's taste as a preference vector and conditions a deep
enhancement network on this representation~\cite{kim2020pienet}. SpliNet embeds
reference retouchers in a user space and predicts global neural-spline color
transforms~\cite{bianco2020personalized}. Masked style modeling further makes a
user's preferred style content-aware and trains with synthetically constructed
input--retouch pairs~\cite{kosugi2024personalized}. A recent Transformer
approach infers a preference vector from a few user-selected images and
conditions enhancement on global and local style information
\cite{kim2025global}. These methods demonstrate
that a single generic output cannot represent the diversity of individual
taste.

Most closely, Personalized Photographic Style (PPS) learning infers a user's
photographic style from pairwise judgments and evaluates adaptations of
style-transfer and enhancement models~\cite{kim2026pps}. PerTouch uses semantic
parameter maps and a VLM-driven agent with feedback-driven rethinking and
scene-aware memory to translate language instructions and feedback into
fine-grained retouching controls~\cite{chang2026pertouch}. These methods
establish nearby preference-to-transformation settings. PPS uses comparative
judgments, while PerTouch uses language instructions and interactive feedback.
Neither published formulation targets federated raw-data-local scorer learning
or unpaired on-device enhancer adaptation.

The supervision and privacy setting of FedPAIE are different. Prior
personalized enhancers are generally trained in a centralized setting from
explicit user choices, preferred-style examples, paired retouches, or
pseudo-paired transformations. FedPAIE does not require a personalized target
image for any photograph in the local enhancement set. A new user provides a
small private rated support set to calibrate the scorer, while the enhancer is
adapted on a separate set of ordinary unpaired local photographs. Generic
paired retouches are used only to learn the common CLUT prior and do not encode
the target style of the new user. This decomposition replaces user-specific
retouch supervision with scorer-mediated preference supervision while keeping
the user's raw photographs and ratings local.

\begin{table*}[t]
\centering
\scriptsize
\setlength{\tabcolsep}{2.5pt}
\renewcommand{\arraystretch}{1.16}
\begin{tabular}{@{}
>{\raggedright\arraybackslash}p{0.135\textwidth}
>{\raggedright\arraybackslash}p{0.145\textwidth}
>{\raggedright\arraybackslash}p{0.18\textwidth}
>{\centering\arraybackslash}p{0.085\textwidth}
>{\centering\arraybackslash}p{0.10\textwidth}
>{\centering\arraybackslash}p{0.095\textwidth}
>{\raggedright\arraybackslash}p{0.19\textwidth}@{}}
\toprule
Method &
Output / inference &
Preference signal &
Fed.\ training &
Raw-data locality &
Pers.\ transform &
Transform supervision / local adaptation \\
\midrule
Federated PIAA~\cite{xiong2023federated} &
Personalized aesthetic score &
Decentralized user image--rating data &
Yes &
Yes (reported) &
No &
N/A \\
\midrule
PAA+~\cite{zhong2025paaplus} &
Personalized aesthetic score &
Surveys, scores, and accumulated multimodal feedback &
Not addressed &
Not reported &
No &
N/A \\
\midrule
PPS~\cite{kim2026pps} &
Personalized photographic-style output &
Pairwise user judgments over style candidates &
Not addressed &
Not reported &
Yes &
Pairwise style candidates\newline
Unpaired local adaptation not reported \\
\midrule
\textbf{FedPAIE} &
\textbf{Personalized enhanced image}\newline
\textbf{0.293M enhancer at inference} &
\textbf{Sparse scalar ratings and unpaired local photos} &
\textbf{Yes (scorer)} &
\textbf{Yes (protocol)} &
\textbf{Yes} &
\textbf{No user-specific retouch target}\newline
\textbf{Unpaired local photos} \\
\bottomrule
\end{tabular}
\caption{Method-level comparison of the closest personalized-aesthetics
settings. ``Not reported'' means that the cited work does not state the
corresponding raw-data-locality or local-adaptation property. ``Not addressed''
marks a capability outside its stated task, and N/A marks a
transformation-specific field that does not apply to a score-only method.
Generic and LUT-based enhancement methods remain complementary architecture
context in Sec.~I.1 rather than being mixed into this task-setting comparison.}
\label{tab:app_novelty_positioning}
\end{table*}

\subsection{Federated Preference and Visual Learning}
\label{app:rw_federated}

Federated learning trains a shared model through client-side optimization and
server aggregation without collecting raw client samples
\cite{mcmahan2017communication}. Extensions address statistical heterogeneity
and non-IID client distributions~\cite{li2020federated,mohri2019agnostic}.
Federated collaborative filtering, matrix factorization, and recommendation
learn user or item representations from decentralized interactions
\cite{ammad2019federated,chai2021secure,liang2021fedrec,yi2021efficient,
liu2023federated}. Their outputs support ranking and recommendation rather than
image formation.

Federated vision research has demonstrated decentralized learning for object
detection, general computer-vision tasks, personalized aesthetic assessment,
and parameter-efficient video moderation
\cite{liu2020fedvision,he2021fedcv,xiong2023federated,
tao2025fedvideomae}. The closest task to FedPAIE is federated personalized
image-aesthetics assessment~\cite{xiong2023federated}, which predicts
user-specific aesthetic scores from decentralized data. As an assessment
method, its output remains a prediction rather than a user-specific enhanced
image. The other federated vision systems demonstrate decentralized visual
learning across deployed, benchmark, and parameter-efficient settings, but do
not learn an individual's color-grading preference or translate it into a
personalized photo transformation.

FedPAIE crosses this task boundary. Federated training supplies a shared
aesthetic scorer initialization, local support ratings calibrate it to a new
user, and the frozen personalized scorer subsequently guides a local enhancer
on unpaired images. Only scorer parameters and scalar aggregation counts
participate in the federated exchange. The generic enhancer is learned
independently from generic paired retouches, and user-specific enhancer
adaptation remains on device. This division limits the federated component to
preference learning instead of communicating a full image-to-image model. As
stated in the main paper, this design preserves raw-data locality by keeping
photos, ratings, and personalized models on device while communicating only
lightweight scorer updates and scalar aggregation counts.

\subsection{Positioning and Technical Novelty of FedPAIE}
\label{app:rw_positioning}

Tab.~\ref{tab:app_novelty_positioning} summarizes the functional boundary
between the closest personalized-aesthetics task settings. FedPAIE introduces an end-to-end
federated personalization formulation that connects Federated Aesthetic
Preference Learning, Personalized Scorer Calibration, and
Frozen-Scorer-Guided Enhancer Adaptation. This formulation directly converts
decentralized sparse ratings into user-specific 3D-LUT transformations on
unpaired local images.
\par\noindent\textbf{From Decentralized Ratings to Image Transformations.}
Previous federated preference methods terminate at a score, ranking, or
recommendation, while personalized enhancement methods directly learn a
transformation from centrally available preference or retouch supervision.
FedPAIE connects these two endpoints through a calibrated scorer that is both
user-specific and differentiable. This connection enables sparse ratings to
guide an enhancer even when the local photographs have no corresponding
personalized targets.
\par\noindent\textbf{Separation of Shared Priors and Private Adaptation.}
FedPAIE learns two independent shared initializations. The aesthetic scorer is
federated across decentralized ratings, while the Generic Enhancement Prior is
initialized from a pretrained checkpoint obtained from ordinary paired
retouches. Personalization then occurs entirely on device
in a fixed order: calibrate the scorer from the private rated support set,
freeze it, and adapt the enhancer from unpaired photographs. This separation
keeps the user's preference evidence local and avoids treating a population
retouching style as the user's target style.
\par\noindent\textbf{Support-Aware and Stability-Aware Personalization.}
The Support-Dependent Scorer Mask restricts the adaptable parameter blocks when
only 10 ratings are available and permits broader calibration with 100 ratings.
The local ranking and variance objectives complement regression by preserving
relative preference information and score dispersion. During enhancer
adaptation, the scorer remains fixed and the preference objective is balanced
by pixel, perceptual, and excess-gap regularization. The resulting design
addresses two distinct risks of few-shot scorer guidance: overfitting the
preference model and over-optimizing its imperfect output.

\par\noindent\textbf{A Lightweight Deployment Boundary.}
Personalization updates only the CNN backbone and coefficient head of the
enhancer while keeping the compressed LUT bases fixed. The scorer is a
training-time component and is removed after adaptation. Inference therefore
retains only the 0.293M-parameter personalized enhancer, which distinguishes
FedPAIE from scorer-only federated aesthetics methods and from personalization
pipelines that retain a larger preference or style model at deployment.

Taken together, these properties establish FedPAIE as a distinct federated
personalization framework. To the best of our knowledge, FedPAIE is the first
federated method for personalized aesthetic image enhancement and color
grading. It transforms decentralized sparse ratings into lightweight
user-specific color grading on unpaired local images while keeping raw photos
and ratings on device.

\end{document}